\documentclass[letterpaper, 10 pt, conference]{ieeeconf}  % Comment this line out if you need a4paper

\IEEEoverridecommandlockouts    
\usepackage{graphicx} % for pdf, bitmapped graphics files
\usepackage{xcolor}
\definecolor{cvprblue}{rgb}{0.21,0.49,0.74}
\definecolor{linkpurple}{HTML}{880ED4} % 6C0AB9
\usepackage[pagebackref,breaklinks,colorlinks,allcolors=linkpurple]{hyperref}

\usepackage{eso-pic}

\title{\LARGE \bf
BLANKET: Anonymizing Faces in Infant Video Recordings
}

\author{Ditmar Hadera, Jan Cech, Miroslav Purkrabek and Matej Hoffmann %
\thanks{The authors are with the Department of Cybernetics, Faculty of Electrical Engineering, Czech Technical University in Prague
        {\tt\small matej.hoffmann@fel.cvut.cz}}%
\thanks{This work was supported by the Czech Science Foundation (GA CR), project no. 25-18113S, EC Digital Europe Programme project CEDMO 2.0 no. 101158609,	and by the CTU grant SGS23/173/OHK3/3T/13.}% <-this % stops a space
}

\begin{document}

% ---- top watermark block for arXiv (first page only) ----
\AddToShipoutPictureFG*{%
  \AtPageUpperLeft{%
    \raisebox{-1.1cm}{% <-- move down/up (increase to move down)
      \makebox[\paperwidth]{%
        \begin{minipage}{0.92\paperwidth}
          \centering
          \color{gray}
          % \rule{\linewidth}{0.4pt}\vspace{2pt}

          {\fontsize{7}{8}\selectfont
          This is the authors' final version of the manuscript published as:\\
          Hadera, Ditmar, et al. "BLANKET: Anonymizing Faces in Infant Video Recordings." 2025 IEEE International Conference on Development and Learning (ICDL) 2025, \textcopyright~IEEE,\\
          \texttt{https://doi.org/10.1109/ICDL63968.2025.11204388}\\
          Code and accompanying video available at \texttt{https://github.com/ctu-vras/blanket-infant-face-anonym}
          \par}

          % \vspace{2pt}\rule{\linewidth}{0.4pt}
        \end{minipage}%
      }%
    }%
  }%
}
% ----------------------------------------------

\maketitle
\thispagestyle{empty}
\pagestyle{empty}

\begin{abstract}
Ensuring the ethical use of video data involving human subjects, particularly infants, requires robust anonymization methods. We propose BLANKET (Baby-face Landmark-preserving ANonymization with Keypoint dEtection consisTency), a novel approach designed to anonymize infant faces in video recordings while preserving essential facial attributes.  Our method comprises two stages. First, a new random face, compatible with the original identity, is generated via inpainting using a diffusion model. Second, the new identity is seamlessly incorporated into each video frame through temporally consistent face swapping with authentic expression transfer. The method is evaluated on a dataset of short video recordings of babies and is compared to the popular anonymization method, DeepPrivacy2. Key metrics assessed include the level of de-identification, preservation of facial attributes, impact on human pose estimation (as an example of a downstream task), and presence of artifacts. Both methods alter the identity, and our method outperforms DeepPrivacy2 in all other respects. The code is available at \url{ https://github.com/ctu-vras/blanket-infant-face-anonym}.
%Sharing datasets including raw video footage is key to make progress in many disciplines, but the identity of the persons on images or videos needs to be protected. Here we focus on the face which carries the most identifying information. Simple methods like covering the face degrade the performance of downstream tasks like head or pose estimation. State-of-the-art computer vision methods are capable of ``swapping'' faces in images of adults. We compare their performance on videos with infants and present a pipeline that changes the face throughout the whole video, preserving temporal consistency as well as performance in downstream tasks like pose estimation.
\end{abstract}

%Ensuring the ethical use of video data involving human subjects, particularly infants, necessitates robust anonymization methods. We propose BLANKET (Baby-face Landmark-preserving ANonymization with Keypoint dEtection consisTency), a novel approach designed to anonymize infant faces in video recordings while preserving essential facial attributes.  Our method comprises two stages: First, a new random face compatible with the original identity is generated through inpainting, employing the diffusion-based Stable Diffusion model. Second, the new identity is seamlessly injected into each video frame using the FaceFusion framework, achieving temporally consistent face swapping with authetntic expression transfer. The method is evaluated on short video reordings of babies and is compared to a popular anonymization method DeepPrivacy2. We measured level of de-identification, preservation of facial attributes, impact on human pose estimation as an example of downstream task, and presence of artifacts. Both method alters the identity, and our BLANKET outperforms DeepPrivacy2 in all other respects. 

% --------------------------------------------------------------------------------

\section{Introduction}
Sharing datasets of human subjects, including video footage, is crucial for advancing psychological science \cite{adolph2017video}. However, for ethical reasons, this cannot be done without anonymizing the videos. To study development, daily spontaneous recordings of infants will be needed \cite{adolph2011sampling} to uncover developmental trajectories at multiple and nested time scales. The issues related to privacy and ethics are even more pressing when it comes to infants and children.  
%Sharing datasets including raw video footage is key to make progress in psychological science. For ethical reasons this is not possible without anonymizing the videos. Simple methods like covering the face degrade the performance of downstream tasks like head or pose estimation.

Anonymization or de-identification is a process of transforming an image that contains identifiable information, such as a face, into another image that preserves the appearance and the context, but hides the identity of the original image. There are trivial ways to hide the identity of a subject like face blurring or masking by a black box. Besides unpleasant and unnatural results, a significant drawback is that any facial information (age, gender, race, gaze, facial expression) is lost together with the identity. Moreover, these manipulations with the original image may harm downstream processes of image analysis, such as pose estimation.

\begin{figure}
    \centering
    \includegraphics[width=0.9\linewidth]{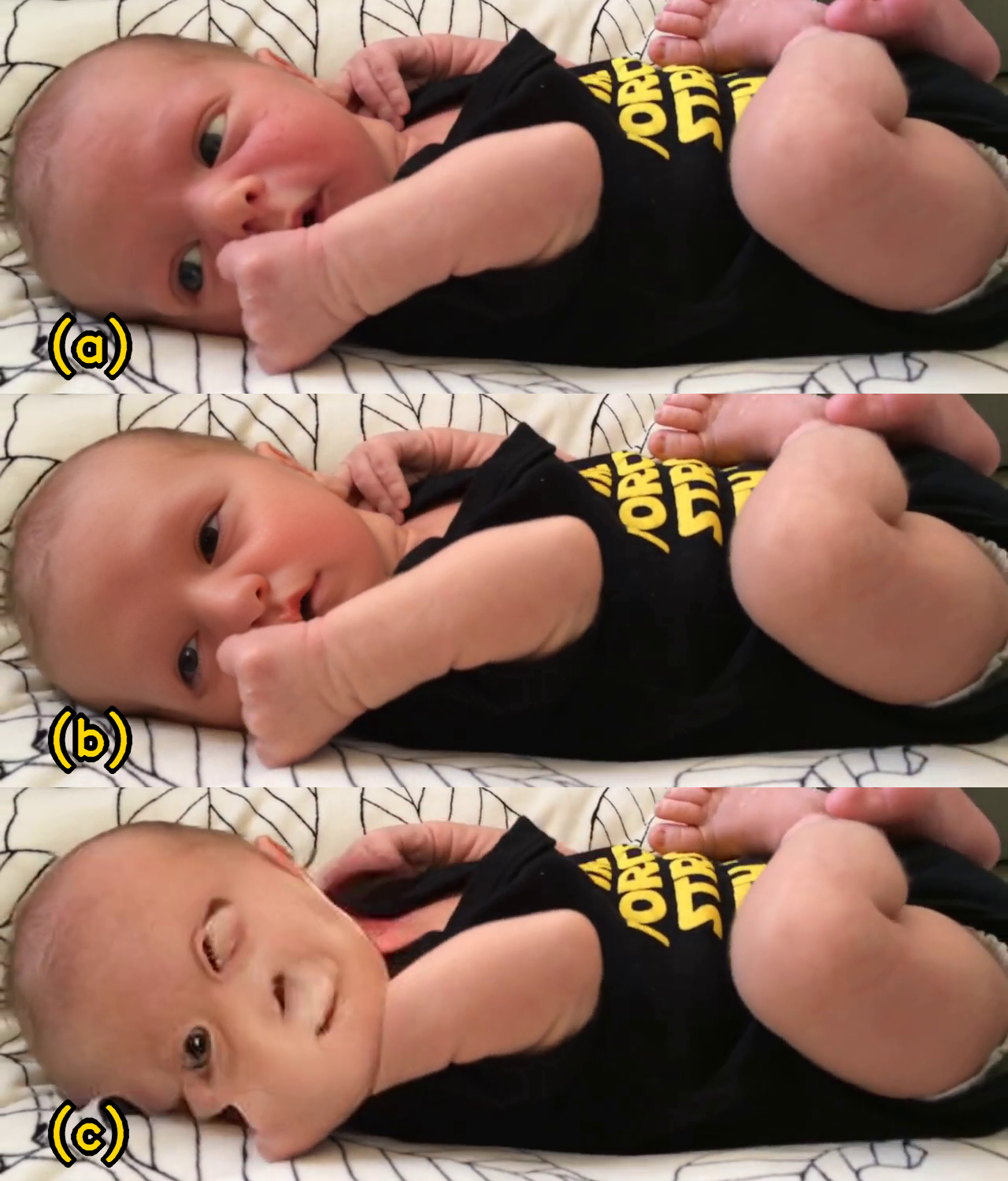}
    \caption{Infant face anonymization. Frame of the original video (a), anonymization by the proposed BLANKET method (b), and by DeepPrivacy2~\cite{deepprivacy2} (c). The proposed method alters the identity of the infant without introducing obvious perceptual artifacts, while keeping all other facial attributes intact, e.g. face orientation, gaze, expression. The method can handle face occlusion, which is challenging for the competing method. Original video courtesy of Stephen Julia, Max Family Fun.}
    \label{fig:teaser}
\end{figure}

We propose a method tuned for the anonymization of infants in video recordings. The method has two stages. First, a novel random face that is compatible with the original identity is generated by inpainting with a diffusion model. Then, the novel identity is injected into every frame of the video using a temporally consistent face swapping. 

The challenge is that the subjects are infants or newborns. This demographic group is underrepresented in computer vision datasets and models typically struggle to generalize to unseen data. Infants have different facial features and proportions than adults. For instance, popular generative models for face synthesis, e.g. StyleGAN2~\cite{Karras2019stylegan2} has difficulties to generate infants, since it was trained on the FFHQ dataset containing mostly photos of adult faces downloaded from Flicker. Therefore, we use, Stable Diffusion~\cite{stable-diffusion}, a large scale generative model to do the inpainting in the first stage. This model was trained on a LAION dataset of billion image scale which certainly includes infants, and so produces an image with a new undistorted identity.

The new identity is inserted into every frame of the video by FaceFusion~\cite{Facefusion}, which ensures efficient, seamless and temporally consistent face swap, with high-fidelity expression transfer from the original recording. 
See Fig.~\ref{fig:teaser} for an example of anonymization performed by our method and by a recent popular method DeepPrivacy2~\cite{deepprivacy2}, a GAN-based model. 

To summarize, our contributions are as follows:
\begin{itemize}
\item A novel method, called BLANKET (Baby-face Landmark-preserving ANonymization with Keypoint dEtection consisTency), was proposed. It is tuned for the anonymization of infants. It possess the following properties: (1) Seamlessly changes the identity of a subject. (2) Preserves facial expressions and all other attributes and states of the face. (3) Produces minimal perceptual artifacts. (4) Is temporally consistent for a video. 
%\item \red{We collected a dataset of about 50 videos of infants downloaded from YouTube.} 
\item The method was extensively evaluated to quantitatively assess its properties, using statistics derived from computer vision estimators and by a user study. We tested the impact of our anonymization to human pose estimation, i.e., finding keypoints on the body and face in an image, as a prominent example of a downstream task.
\item The code is available at \url{ https://github.com/ctu-vras/blanket-infant-face-anonym}.
\end{itemize}

The remainder of the paper is structured as follows. Related work is reviewed in Sec.~\ref{sec:related-work}. The method is presented in Sec.~\ref{sec:method}. Experiments are described in Sec.~\ref{sec:experiments}. We close with Conclusion, Discussion, and Future Work in Sec.~\ref{sec:conclusion}.

%Challenges
% - subjects are infants
% - video consistency

%Contributions:
% - metoda
% - extensivni vyhodnoceni, vcetne vlivu na odhad pozy a user study
% - collected dataset Youtube-baby

% --------------------------------------------------------------------------------

\section{Related Work} \label{sec:related-work}

There exists a wide variety of anonymization techniques to conceal the identity of a person in images or video recordings. We refer to a survey~\cite{overview} or a recent journal article~\cite{kim2025diffface} for a more detailed taxonomy of the methods. There are two approaches: Face obscuring and Face modifying methods. % all differ mainly in two aspects: (i) how well they anonymize the face, i.e. how hard it is to figure out who the person in the image was before the anonymization method has been applied; (ii) how much the anonymized face still looks like a human face.

\subsection{Face Obscuring Methods}

These methods are based on obfuscating a face or its part. They include blurring, pixelization, or masking. % A few images of faces anonymized using these techniques can be seen in \figref[face-obscuring-images]. 
% simpler methods
% methods like putting a black square over the face, blurring the face, putting a bar over the eyes, method used in V síti
% add images depicting different simple anonymization mathods used on the same image
The advantage of these methods is their simplicity. It allows them to be fast and not resource intensive. They rely on face detection and tracking and a trivial operation in the face regions. On the other hand, their disadvantage is that their de-identification effectiveness is directly proportional to how much the anonymized face still looks like a real face. Using images generated with these methods on downstream tasks is therefore impossible most of the time. For instance, placing a black box over a subject's face effectively conceals their identity, but head rotation, gaze, or facial expressions are lost completely. 

\subsection{Face Modifying Methods}
Face modification methods focus on seamlessly altering identity while ideally retaining facial attributes such as expressions, gaze, and pose. The face is typically replaced by another real or synthetic face. These methods ensure high realism and utility for downstream tasks such as human behavior analysis. These methods are usually more computationally intensive than the obscuring methods. 

%This allows them to anonymize their identities while still keeping all the aspects of a human. This level of realism makes them usable for anonymizing data and not loosing the possibility of using it for downstream tasks. Unfortunately, it also makes them more complex, which results in higher resource intensivness and longer anonymization times.

%The existing methods can be distinguished based on whether they utilize deep learning techniques or not.

Before deep learning, the methods utilized face-swap by finding a matching face of a different identity in a large dataset. The new face replaced the original face and was finally blended into the original image. Examples of these methods are the k-Same algorithm~\cite{k-same} and AnonySwap~\cite{anonyswap}.

%The non-deep-learning methods mostly utilize face swapping techniques. The face that is being anonymized is compared to many faces from a large dataset. Then one or multiple of the faces that are the closest by some metric are picked. The image is then anonymized through interpolation. Examples of these methods are the k-Same algorithm \cite{k-same} and AnonySwap \cite{anonyswap}.

More recent methods often employ deep generative models for anonymization. The methods are typically based on inpainting~\cite{sun2018natural}, or face swapping~\cite{natsume2019fsnet}. The generative models used are based on Generative Adversarial Networks (GANs)~\cite{wu2019privacy,maximov2020ciagan,deepprivacy, deepprivacy2}, or more recently on Diffusion Models~\cite{zhao2023diffswap,kim2025diffface}.

 Consistent preservation of the above-mentioned facial properties is usually not guaranteed. Moreover, all the above methods work for images and not for a video recording. Unpleasant flickering artifacts would appear if used naively frame-by-frame on a video. The problem of video anonymization seems to be unexplored, although there are exceptions~\cite{balaji2021temporally}. A recent paper~\cite{rosberg2023fiva} reports promising results in videos. However, the code is not publicly available, so prospects on infant face anonymization remain unclear.

%The methods utilizing deep learning approach the anonymization in multiple different ways. Examples of these methods include StyleGAN3 \cite{stylegan3}, or DeepPrivacy 2 \cite{deepprivacy2}.

% changing the faces in a way where they still look like faces but the identity is different
% possible approaches:
%   1. face swapping using a large dataset
%   2. image inpainting

% Moravčík, DeepPrivacy 1 and 2,...

% they allow to anonymize but the resulting face still looks like a face
% there is a connection between the effectiness of anonymization and preservation of facial expression

% --------------------------------------------------------------------------------

\section{Method} \label{sec:method}

%Zakladni myslenka - pouzijeme pokrocilou metodu faceswap, facefusion. Vstupem je puvodni video a obrazek jako nova identita. Metoda zajisti ze vystupni video bude obsahovat novou identitu, bez prostorovych a casovych artifaktu. 

%Potrebujeme novou identitu, ktera bude dostatecne jina nez ta puvodni, ale zaroven kompatibilni. Neni napr. mozne nasledny faceswap provest s oblicejem dospele oosby na novorozence. Novy oblicej by mel byt stejne rasy a pohlavi, s priblizne podobnymi geometrickymi proporcemi. Proto generovani nove identity - nahodne pomoci inpainingu s difusnim modelem Stable Diffusion. 

%Nevím jak moc pravdivé je, že naše metoda seamlessly prohodí originální identitu s novou bez švů, problémů s osvětlením a různým blikání. Některá vygenerovaná videa mají relativně viditelné švy, nasvícení taky asi není vždy úplně konzistentní (často se tváře tak jakoby lesknou) a některá videa také mají problém s problikáváním tváře (např. 000017).

\begin{figure}
\centering
\includegraphics[width=1\linewidth]{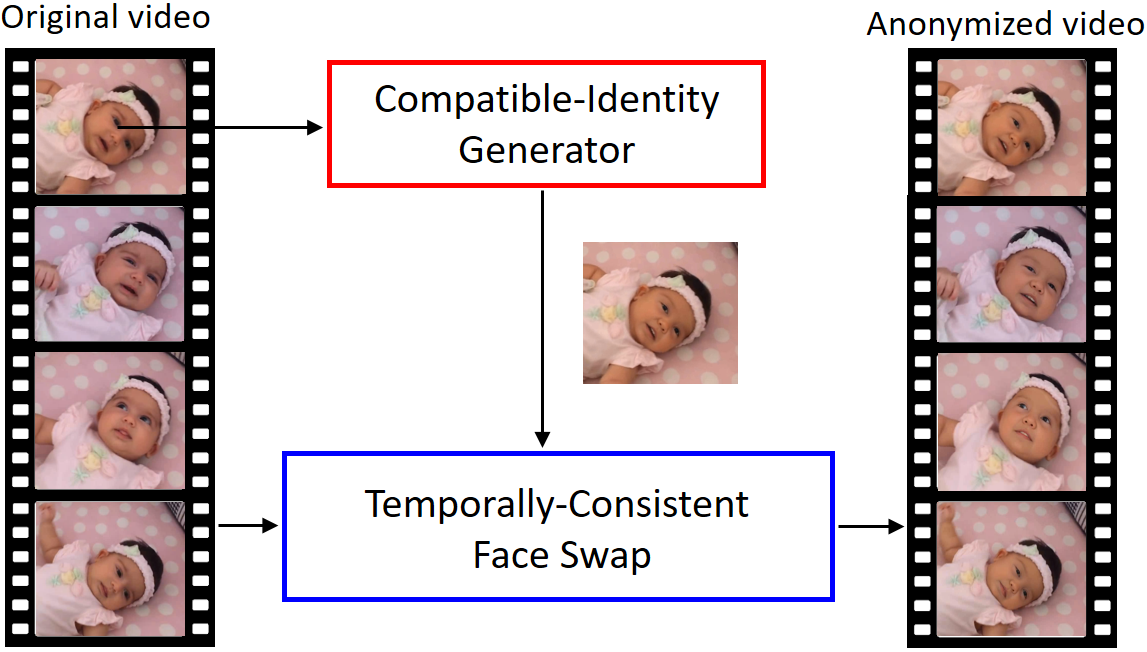}
\caption{Flowchart of the proposed video anonymization. First, an image replacing the original identity with a new compatible random identity from the first frame is created. The new face is generated by inpainting using Stable Diffusion~\cite{stable-diffusion}. Then, the new identity is swapped in every frame of the video. We propose to use FaceFusion~\cite{Facefusion}, as it provides temporally consistent results while preserving original facial expressions. }
\label{fig:flowchart}
\end{figure}

Here we propose BLANKET (Baby-face Landmark-preserving ANonymization with Keypoint dEtection consisTency). The fundamental idea of this anonymization method is to employ a temporally consistent face swapping algorithm. The input consists of an original video and an image representing the new identity; see Fig.~\ref{fig:flowchart}. This technique ensures that the original identity is seamlessly replaced with the new identity in the output video, avoiding spatial artifacts (e.g., obvious seams, lighting inconsistencies,  or spurious details) and temporal inconsistencies (e.g., blinking or flickering). In addition, except for the change of identity, all other attributes of the face are preserved, including head orientation, gaze, eyeblinks, facial expressions, and emotions. The output anonymized video looks natural for a human and is usable for downstream tasks such as facial analysis and body pose estimation, in the same way as the original video. We propose using FaceFusion~\cite{Facefusion}, which is a recent framework for advanced face swapping. FaceFusion detects, tracks, and aligns faces and performs face swapping. The tool incorporates lip-syncing, expression-matching, and a face enhancement model that together ensure consistency in pose, expression, and lighting. 

The challenge is to find an image of a new identity that is significantly different from the original but still compatible. For example, it is not possible to perform a faceswap with the face of an adult onto a newborn. The new face should be of the same race and gender, with approximately similar geometric proportions. Therefore, inspired by~\cite{hadera-2024}, we propose generating the new identity randomly using inpainting with the diffusion model, namely Stable Diffusion~\cite{stable-diffusion}.

\begin{figure}
    \centering
    \includegraphics[width=0.9\linewidth]{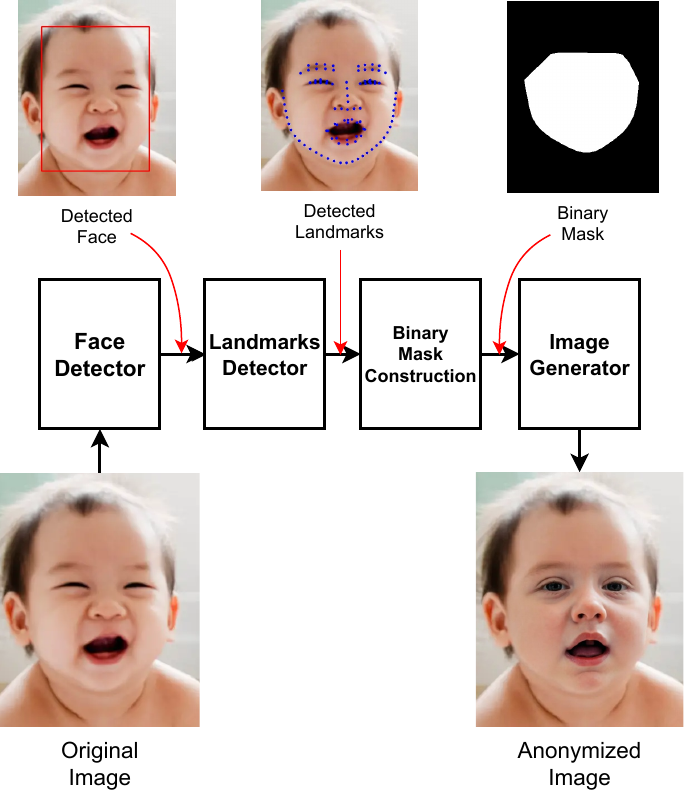}
    \caption{Flowchart of the Compatible-identity generator. A random identity is generated by inpainting algorithm, which makes the new identity compatible, well fitting, and seamlessly merged with the original image.}
    \label{fig:diagram}
\end{figure}

The pipeline that produces the input image of the novel compatible identity is sketched in Fig.~\ref{fig:diagram}. The first frame of the video, where the face is detected, is used. It consists of four steps:
(1) Detecting all faces in an image,
(2) Finding facial landmarks of each detected face,
(3) Generating a binary mask for the detected face, and
(4) Inpainting the masked-out face in the original image to anonymize it.

The face detector provides positions of the faces as rectangular bounding boxes.  We use YOLO11 \cite{yolov11} to detect faces \cite{yolo-face}. %When detecting faces the minimum detection confidence is set to 50\%.
The face bounding boxes are used to find the locations of facial landmarks. The landmarks correspond to important facial features, such as the eyes, mouth, and nose. We use the SPIGA model~\cite{spiga} with ``wflw'' weights, which detects 98 distinct landmarks and predicts a head pose. Then we construct a binary mask, as a convex hull of the landmarks. The mask delineates the inner area of the face to be modified by inpainting. 

Image inpainting, as described in~\cite{inpainting}, is the process of filling in missing parts of an image with such content that an unknowing observer would not notice the modification. Since the inpainting has no or very limited access to the region inside the mask, the inpainting algorithm generates a random identity that matches the area outside of the mask. In this way, an anonymized image is finally generated.  

We employ the generative model called Stable Diffusion \cite{stable-diffusion} as our image generator. %Specifically its implementation called Stable Diffusion WebUI  \cite[stable-diffusion-webui]. 
For generating images, we use two different checkpoints: ``Realistic Vision V2.0''\footnote{\url{https://huggingface.co/SG161222/Realistic_Vision_V2.0}} and ``Realistic Vision V6.0 B1''\footnote{\url{https://civitai.com/models/4201/realistic-vision-v60-b1}}. 

The inpainting algorithm first adds noise to the inpainted area which is subsequently removed by the model. It means that the algorithm does not strictly rely on the context of the face only but uses limited information on the face itself. The level of noise added controls the trade-off between the level of deidentification and the compatibility. We observed that a high level of noise produces visible artifacts more often, while a too low level of noise generates identities too close to the original ones. 
%\red{Pouzivame positivni(jaky?) a negativni promt, CFG, control-net, ...}

We use a positive prompt ``a face of a baby'' and a standard negative prompt. We used CFG~\cite{CFG-ho} and Control-NET~\cite{Control-NET} to fit the landmarks and face geometry and empirically tuned all parameters to balance the above trade-off. The complete parameter setting can be found in the code repository. 

To summarize the method, the first frame of the video is anonymized by generating a random compatible identity by inpainting. Then, this image is fed to the FaceFusion faceswap pipeline as a target identity input together with the original video, which produces the anonymized video. Inpainting with the diffusion model takes about 8 seconds on a consumer GPU but is executed only once, while the FaceFusion that produces the final video runs close to real time. 

% --------------------------------------------------------------------------------

\section{Experiments} \label{sec:experiments}

In the experiments, we extensively evaluate the proposed anonymization method and compare it with a recent popular tool, DeepPrivacy2~\cite{deepprivacy2}. 

First, in Sec.~\ref{sec:stats}, we assess the methods using several statistical measures that reflect the level of de-identification and the preservation of facial attributes.  In Sec.~\ref{sec:pose} we will measure the impact of the anonymization to a downstream task, namely the pose estimation. In Sec~\ref{sec:user_study}, we will present our user study. Finally, Sec.~\ref{sec:qualitative} showcases qualitative results of both methods. 

{For evaluation, we used a subset of Infant Pose Estimation dataset by Chambers et al.~\cite{YTdataset} presented in~\cite{chambers2020computer}. Each of 46 videos of infants is approximately {100 seconds} long, the total footage time is {78 minutes} comprising about {140k} frames.}

%For evaluation, we collected a dataset of 46 videos of infants downloaded from YouTube. Each video is approximately {100 seconds} long, the total footage time is {78 minutes} comprising about {140k} frames. 

%popis statistik - co merime 
%- mira deidentifikace, zachovani atributu, 
%- pro videa: zachovani identity a casova korelace landmarku

%ukaze se, ze nerusi donwstream tasks

%human study (tim se meri oboje, artefakty a mira deidentifikace)

%qualitative results

\subsection{Anonymization statistics} \label{sec:stats}

\begin{table}
    \caption{De-identification quality}
    \begin{center}
    \begin{tabular}{|l||c|c|c|}
        \hline
        Metric & BLANKET & DeepPrivacy2 \\ 
        \hline
        %Same Identity Ratio $\downarrow$ & $0.982 \pm 0.095$ & $0.91 \pm 0.18$ \\ 
        Identity Cosine Distance $\uparrow$ & $0.11 \pm 0.18$ & $0.19 \pm 0.26$ \\ 
        \hline
    \end{tabular}
    \label{tab:deidentification-quality}
    \end{center}
\end{table}

To evaluate the level of de-identification, we use standard face recognition model, ArcFace~\cite{ArcFace}. The model outputs $L_2$-normalized descriptors for each face image. Small cosine distance indicates faces of the same identity, while high cosine distance occurs for different identities. Cosine distance $1-\mathbf{v_o}^T\mathbf{v_a}$ between pairs of frames of the original $\mathbf{v_o}$ and anonymized $\mathbf{v_a}$ identity descriptors is calculated. Tab.~\ref{tab:deidentification-quality} shows average cosine distances calculated over frames of all test videos. It is seen that the difference between the models is not very significant, which we hypothesize is because the ArcFace model was not trained and does not generalize well to infants and newborns. So to assess the level of de-identification, we refer to our user study and qualitative results. 

\begin{table}
    \caption{Gender, race and emotion preservation}
    \begin{center}
    \begin{tabular}{|l||c|c|}
        \hline
        Metric & BLANKET & DeepPrivacy2 \\ 
        \hline
        Same Gender Ratio $\uparrow$ & $0.81 \pm 0.21$ & $0.79 \pm 0.16$ \\ 
        Same Race Ratio $\uparrow$ & $0.69 \pm 0.19$ & $0.56 \pm 0.18$ \\ 
        Same Emotion Ratio $\uparrow$ & $0.51 \pm 0.13$ & $0.27 \pm 0.11$ \\ 
        \hline
    \end{tabular}
    \label{tab:gender-race-emotions}
    \end{center}
\end{table}

To assess how much the anonymization methods preserve facial atributes of the original video, we measure several statistics. All facial attributes are estimated by the DeepFace library~\cite{deepface}. Gender (male, female), race (Indian, Asian, Latino Hispanic, Black, Middle Eastern, White) and emotions (sad, angry, surprise, fear, happy, disgust, neutral) are estimated in both original and anonymized videos. Tab.~\ref{tab:gender-race-emotions} shows the ratio of frames in which the classification is the same. It is seen that the proposed method preserves these attributes better, especially for emotions, due to the fidelity of the expression transfer from the original to the anonymized video. 

\begin{table}
    \caption{Gaze, eye and mouth openness}
    \begin{center}
    \begin{tabular}{|l||c|c|}
        \hline
        Metric & BLANKET & DeepPrivacy2 \\ 
        \hline
        Gaze Difference  $\downarrow$ & $0.36 \pm 0.37$ & $0.39 \pm 0.38$ \\ 
        Eye Openness Difference $\downarrow$ & $0.094 \pm 0.098$ & $0.13 \pm 0.12$ \\ 
        Mouth Openness Difference $\downarrow$ & $0.11 \pm 0.15$ & $0.17 \pm 0.18$ \\ 
        \hline
    \end{tabular}
    \label{tab:gaze}
    \end{center}
\end{table}

We further measure preservation of gaze, openness of eyes and mouth, see Tab.~\ref{tab:gaze}. Gaze is a 2D vector that spans the interval $[-1,1] \times [-1,1]$ for the left-right and up-down directions. The gaze difference is calculated as the Eucledian vector difference between the original and anonymized video.  The openness of the eyes and mouth is calculated as the aspect ratio $d_{tb} / d_{lr}$ between the top-bottom and left-right distances between the respective landmarks in the shape outline, following~\cite{soukupova2016eye}. The difference between the original and anonymized frames is calculated and averaged over all frames in the dataset. Our method achieves better preservation of the openness of the eye and mouth. 

\begin{table}
    \caption{Face orientation preservation}
    \begin{center}
    \begin{tabular}{|l||c|c|}
        \hline
        Metric & BLANKET & DeepPrivacy2 \\ 
        \hline
        X-axis Angle difference [rad] $\downarrow$ & $0.14 \pm 0.32$ & $0.45 \pm 0.57$ \\ 
        Y-axis Angle difference [rad] $\downarrow$ & $0.08 \pm 0.11$ & $0.18 \pm 0.16$ \\ 
        Z-axis Angle difference [rad] $\downarrow$ & $0.08 \pm 0.13$ & $0.17 \pm 0.15$ \\ 
        \hline
    \end{tabular}
    \label{tab:orientation}
    \end{center}
\end{table}

The difference between the orientation of the head before $\mathbf{R_o}$ and after anonymization $\mathbf{R_a}$ is calculated as $\Delta \mathbf{R} = \mathbf{R_o}\mathbf{R_a}^T$ and is decomposed to the Euler angles of the rotation matrix. The results in Tab.~\ref{tab:orientation} show that the head orientation is better preserved in our method.  

\begin{table}
    \caption{Video statistics - identity fluctuation, temporal correlation of landmark trajectories}
    \begin{center}
    \begin{tabular}{|l||c|c|}
        \hline
        Metric & BLANKET & DeeppPrivacy2 \\ 
        \hline
        Original Identity Variance & $0.016 \pm 0.031$ & $0.016 \pm 0.031$ \\ 
        Anonymized Identity Variance $\downarrow$ & $0.025 \pm 0.039$ & $0.051 \pm 0.063$ \\ 
        Correlation of landmarks$\uparrow$ & $0.956 \pm 0.064$ & $0.86 \pm 0.14$ \\ 
        \hline
    \end{tabular}
    \label{tab:video_stats}
    \end{center}
\end{table}

Finally, we calculate the following statistics meant for video consistency, see Tab.~\ref{tab:video_stats}. First, we evaluate how much the identity fluctuates over the frames of the video. Ideally the identity should stay constant over the anonymized videos. ArcFace descriptors are evaluated for every frame. 
%Then, the identity variance is $1 - \frac{1}{T} \sum_{t=1}^T \mathbf{v}_t^T \mathbf{v_\mu}$, where $\mathbf v_\mu$ is the mean ArcFace descriptor computed over the video frames $t \in \{ 1, \dots, T\}$.
Then, the identity variance is $\mbox{var} \{ \arccos( \mathbf{v}_t^T \mathbf{v_\mu}) \}$, where $\mathbf v_\mu$ is the $L_2$-normalized median ArcFace descriptor computed over the video frames $t \in \{ 1, \dots, T\}$.
This quantity is computed for the original videos to provide an idea of the natural identity fluctuations. For videos anonymized by our method, the fluctuation is slightly higher, while it is much higher for the DeepPrivacy2. 

The consistency of the expression is measured by temporal cross-correlation of facial landmark trajectories between the original and anonymized videos. Our method achieves higher correlation, which confirms that the expression is well captured. 

\subsection{Impact of anonymization to pose estimation}
\label{sec:pose}

\begin{table}
    \caption{Relative performance in person detection}
    \begin{center}
    \tabcolsep=4pt
    \begin{tabular}{|l||c|c|c|c|}
        \hline
        Anonymization & none & black rectangle & DeepPrivacy2 & BLANKET \\ 
        \hline
        detection AP & 98.1 & 50.9 & 81.5 & 90.7 \\ 
        \hline
    \end{tabular}
    \label{tab:detection_results}
    \end{center}
\end{table}

\begin{table}
    \caption{Relative performance in human pose estimation}
    \begin{center}
    \begin{tabular}{|l||c|c|c|c|}
        \hline
        Anonymization & pose AP & AP w/o face & in-the-wild AP\\ 
        \hline
        none            & 100.0   & 100.0 & 100.0 \\
        black rectangle & 18.1    & 38.4  & 17.3  \\
        DeepPrivacy2  & 79.1    & 92.3  & 72.3 \\
        BLANKET           & 97.2    & 97.7  & 91.7 \\ 
        \hline
    \end{tabular}
    \label{tab:pose_results}
    \end{center}
\end{table}

We evaluate performance degradation in a downstream task: Human Pose Estimation (HPE).
State-of-the-art HPE methods follow a top-down pipeline, where a person detector is applied first, followed by a pose estimator for each detected bounding box.
We use RTMDet-l \cite{RTMDet} for person detection due to its efficiency and competitive accuracy.
For pose estimation, we adopt the current SOTA ViTPose-b \cite{ViTPose}, trained jointly on COCO \cite{COCO}, MPII \cite{MPII}, and AIC \cite{AIC}.

As ground-truth (GT) annotations are not available for our test videos, we measure performance drop relative to the non-anonymized version.
{We sample frames uniformly from 46 videos from the dataset, resulting in 8897 images.}
Following standard top-down HPE practice, we detect persons using RTMDet and estimate pose for boxes with confidence above 0.3 using ViTPose.
Duplicate detections are then filtered using pose-based non-maximum suppression (NMS), which results in 9102 pseudo GT annotations.

Performance is measured using average precision (AP), following the COCO evaluation protocol.
AP is computed globally over all sampled frames, so Tabs. \ref{tab:detection_results} and \ref{tab:pose_results} do not report standard deviation.

Our method is compared to DeepPrivacy2 and placing a black rectangle on the face.
While the black rectangle anonymizes the person well, it destroys all information about the face.
The analysis below shows that covering face destroys not only facial information, but also hinders detection and pose estimation performance.

\begin{figure}[tb]
    \centering
    \includegraphics[width=0.24\linewidth,page=13]{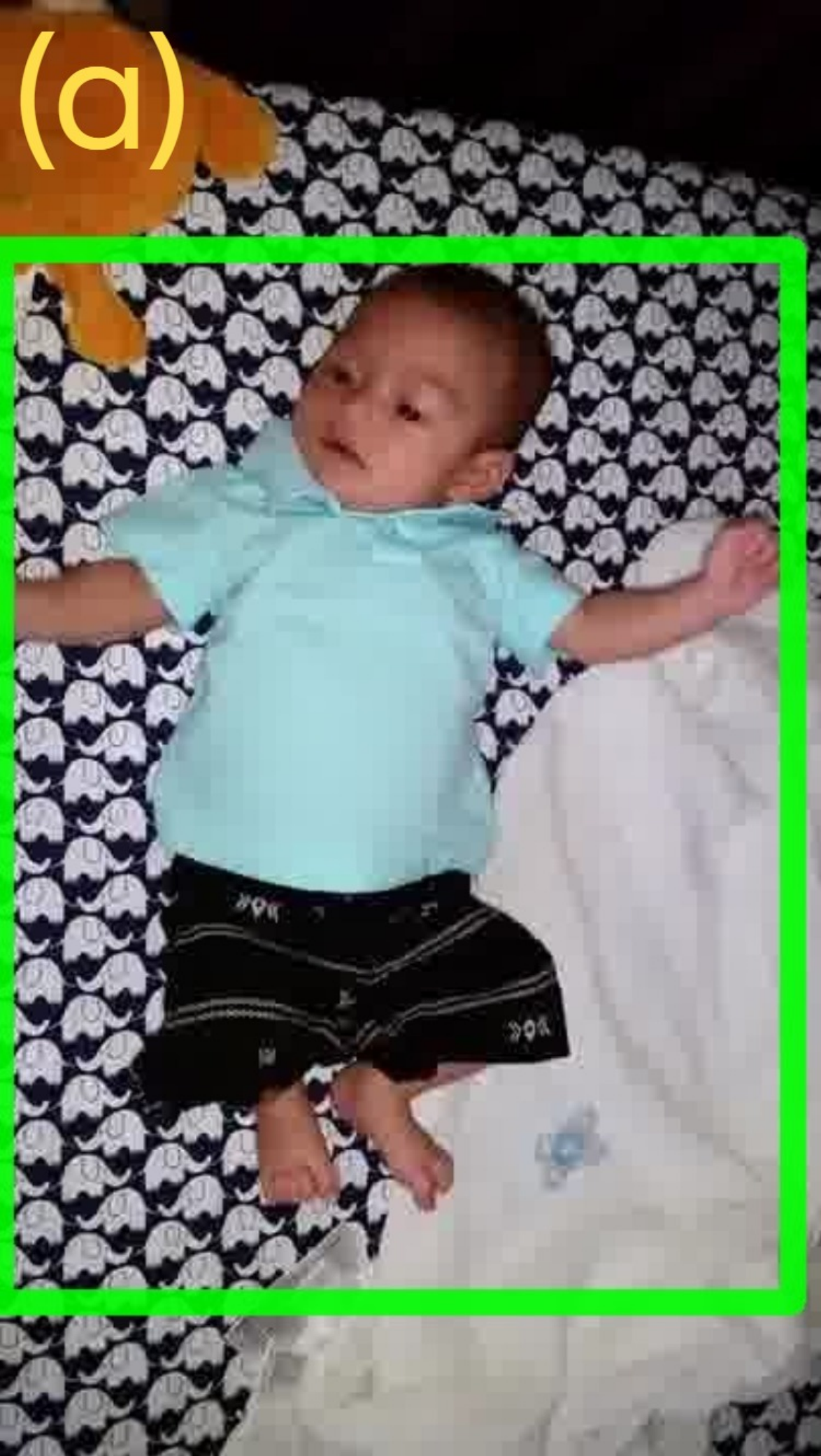}
    \hfill
    \includegraphics[width=0.24\linewidth,page=14]{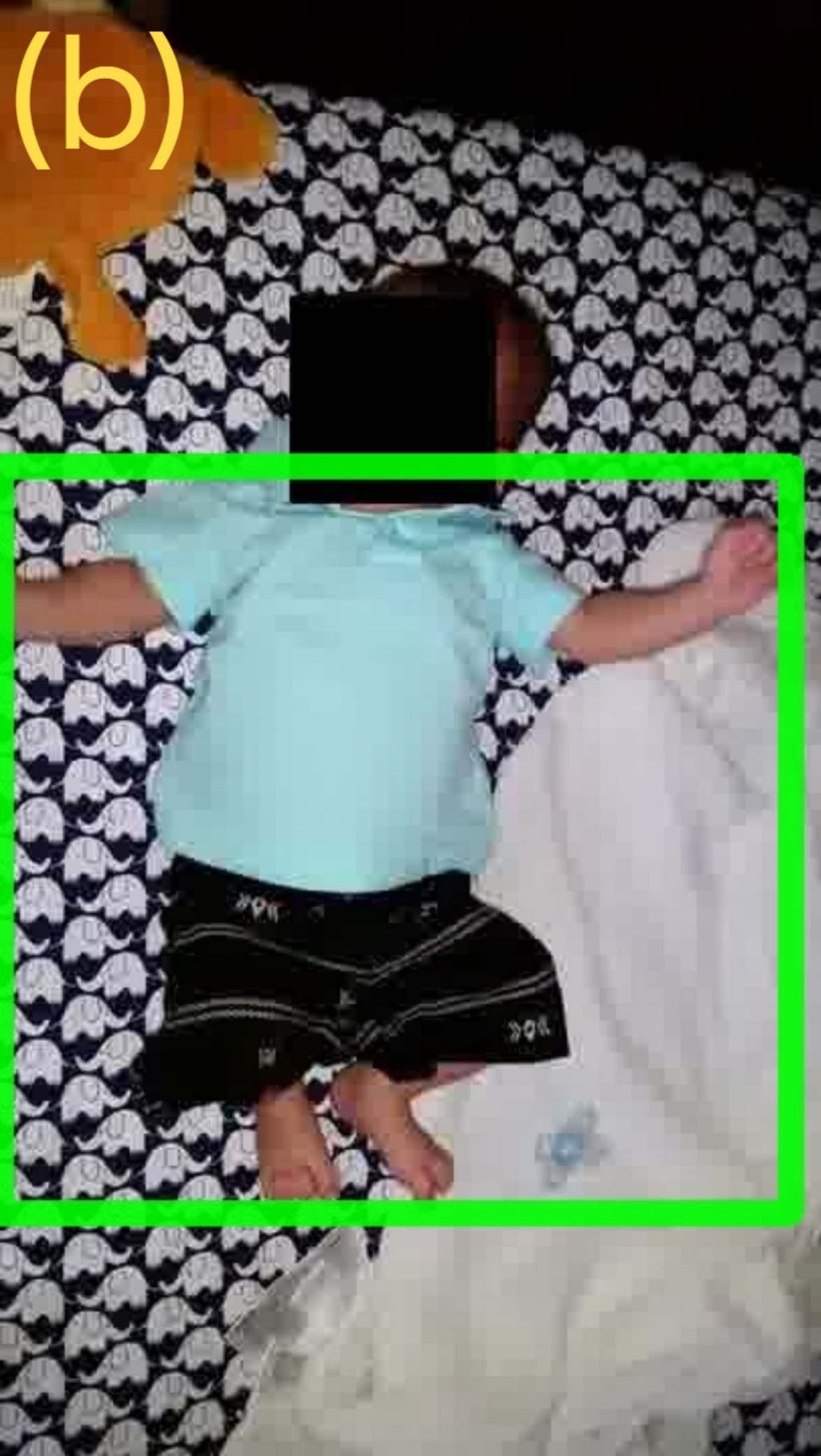}
    \hfill
    \includegraphics[width=0.24\linewidth,page=15]{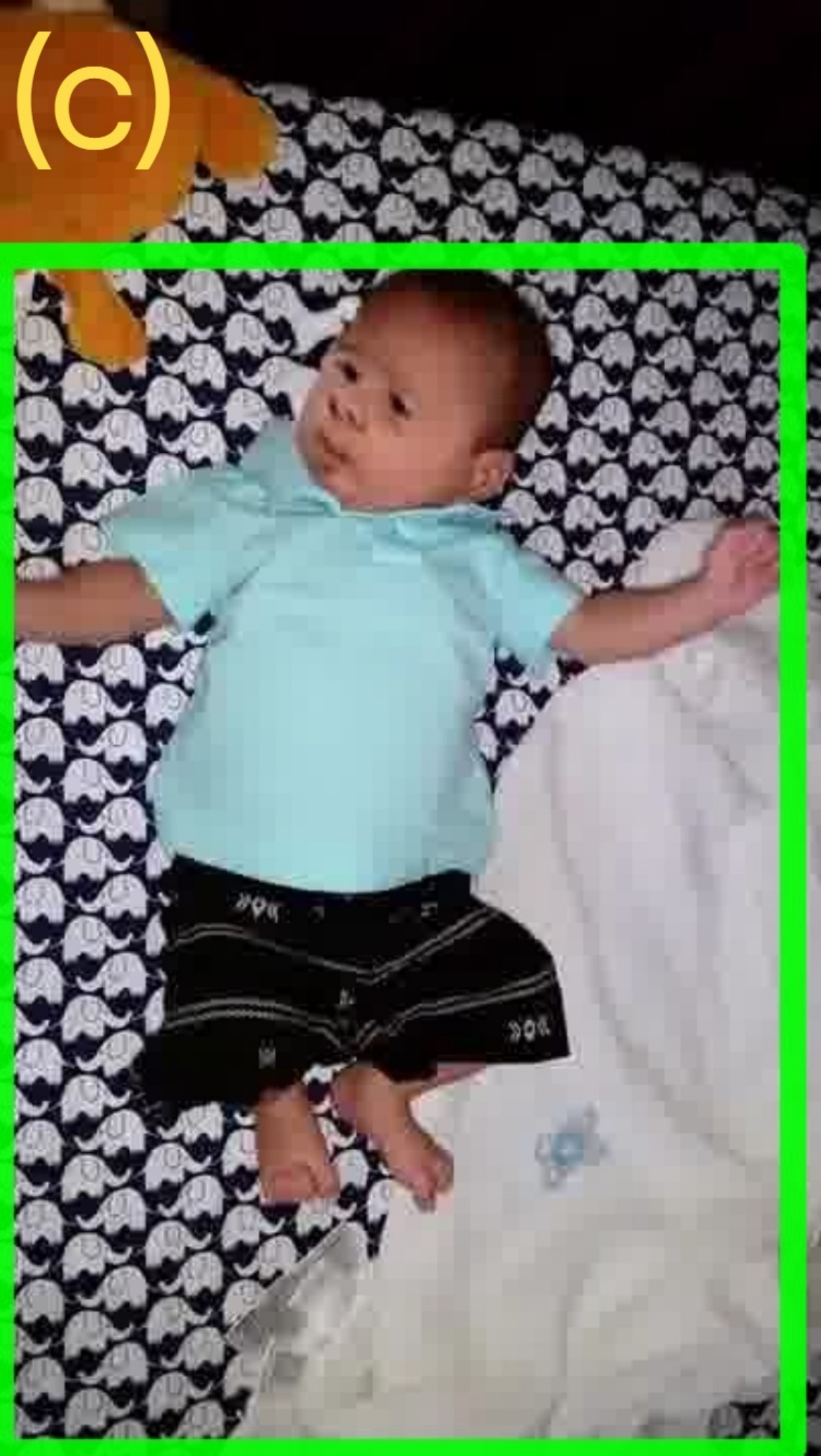}
    \hfill
    \includegraphics[width=0.24\linewidth,page=16]{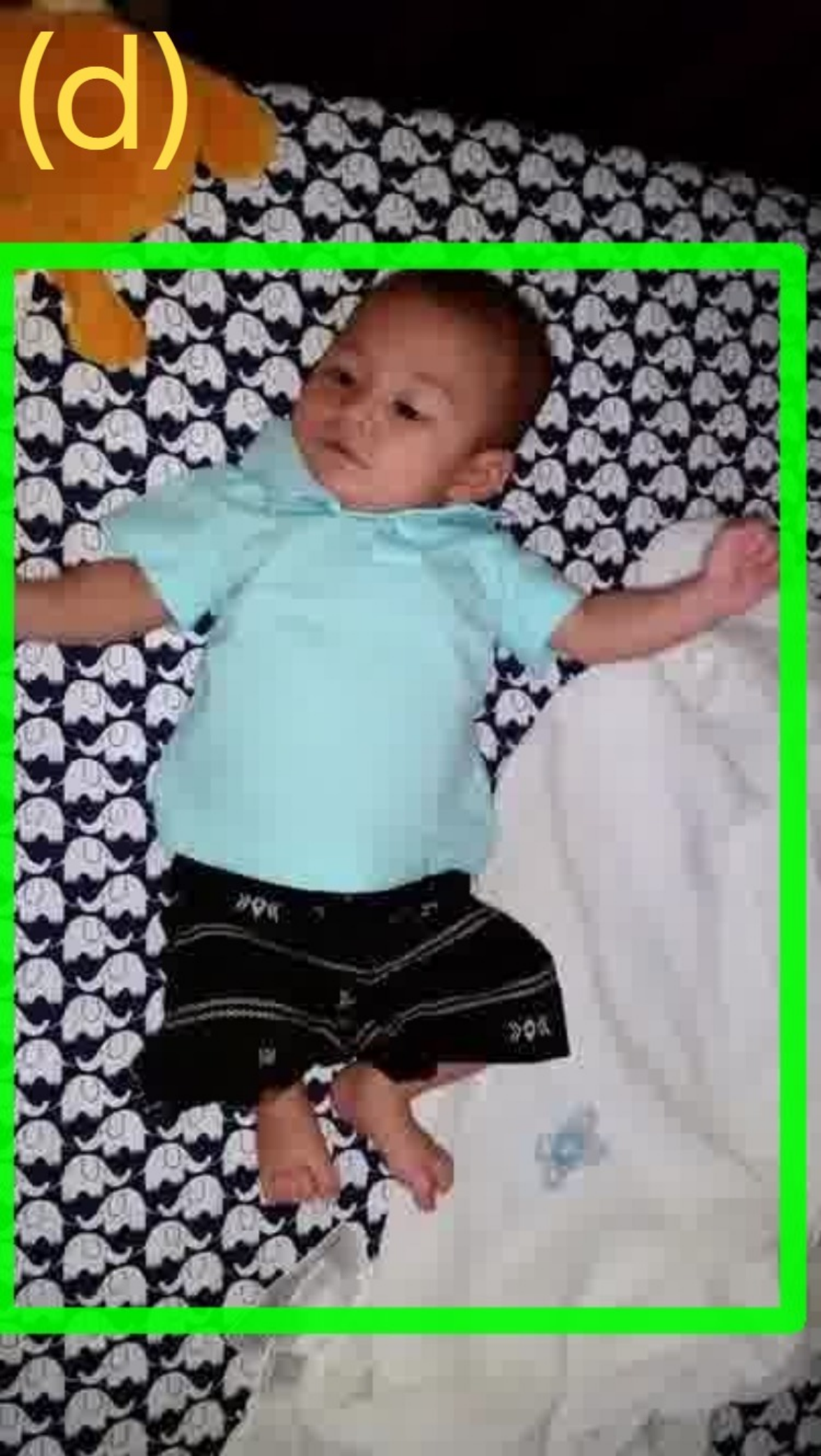}
    \caption{
    % Influence of face anonymization on person detection.
    Anonymization could change output of a detector.
    Original image~(a), black rectangle anonymization~(b), DeepPrivacy2~(c) and BLANKET~(d) with their respective bboxes detected by RTMDet~\cite{RTMDet}.
    Missing face information truncates the head (b) or enlarge detection~(c).
    % Anonymization in (c) enlarged the detected bounding box.
    }
    \label{fig:detection_vis2}
\end{figure}

\begin{figure}[tb]
    \centering
    \includegraphics[width=0.49\linewidth,page=9]{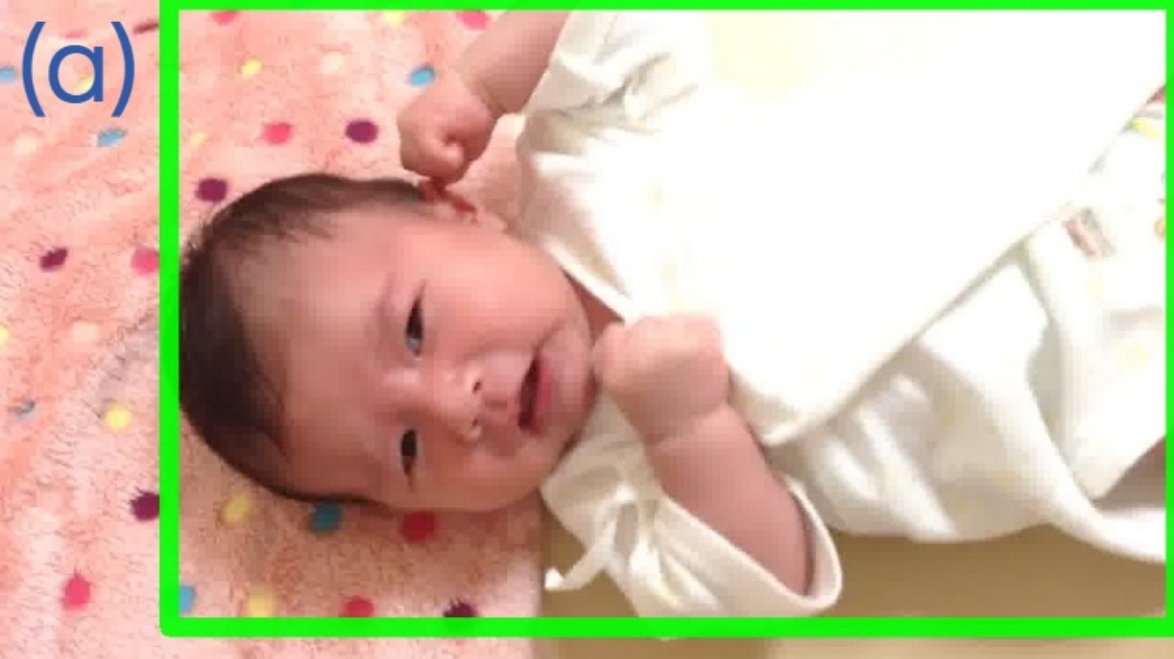}
    \hfill
    \includegraphics[width=0.49\linewidth,page=10]{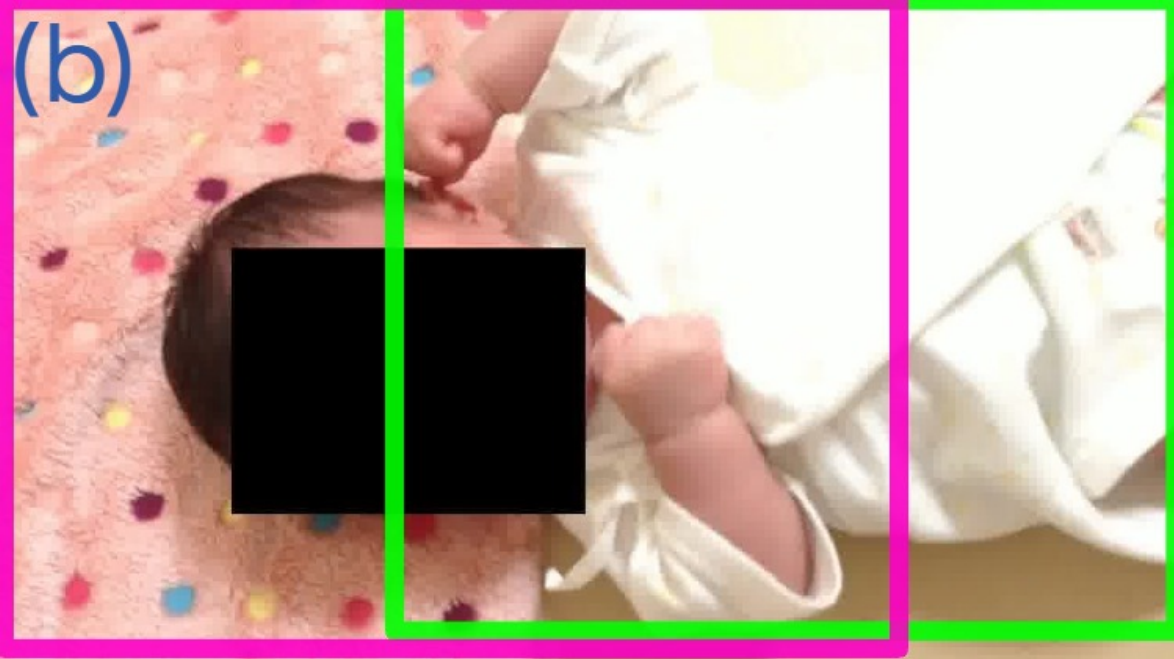}
    \vspace{-1.8ex}
    
    \includegraphics[width=0.49\linewidth,page=12]{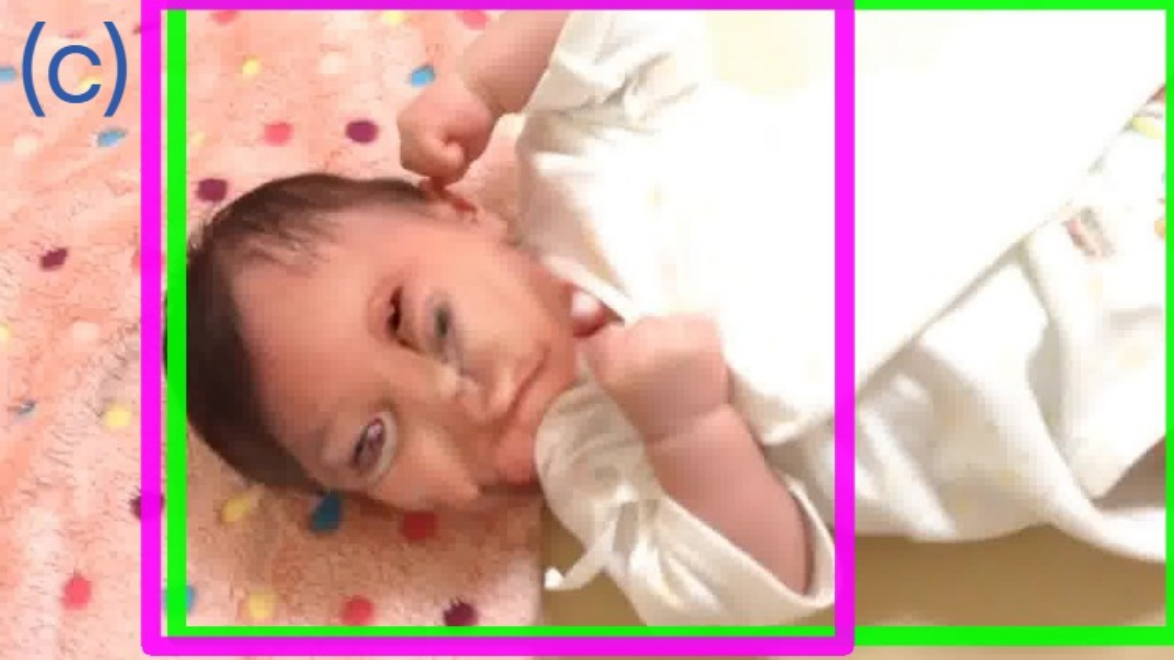}
    \hfill
    \includegraphics[width=0.49\linewidth,page=11]{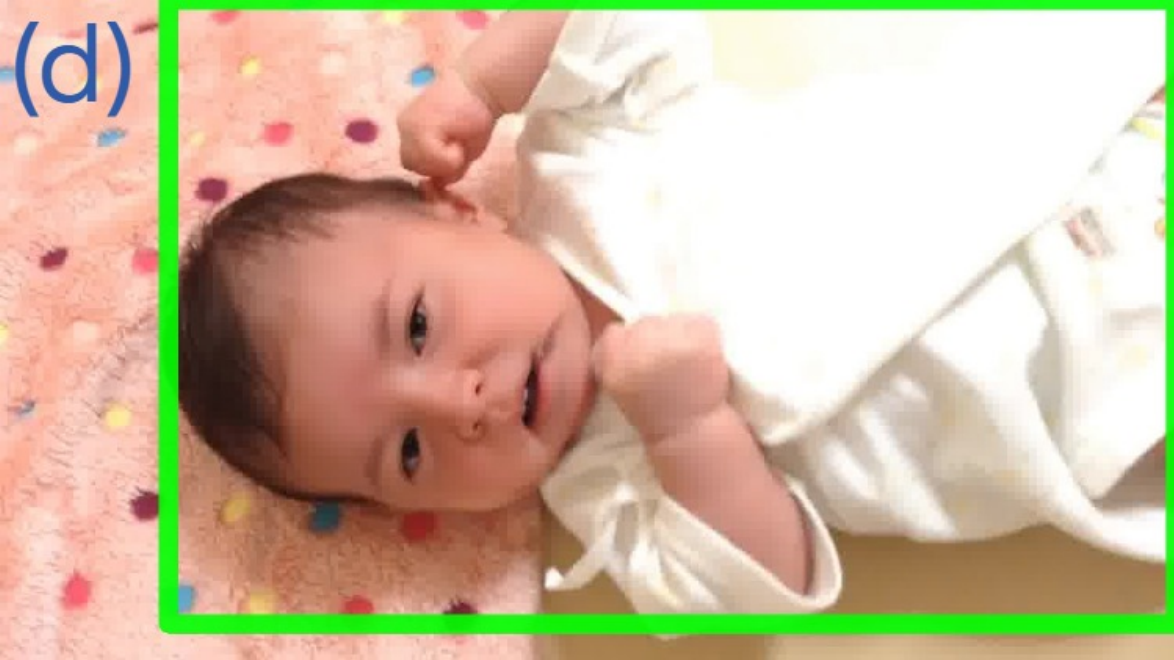}
    \caption{
    % Influence of face anonymization on person detection.
    Deformed face could lead to wrong number of detections.
    Original image (a) with it's bounding box detected by RTMDet \cite{RTMDet}, black rectangle anonymization (b), DeepPrivacy2 (c) and BLANKET (d).
    Missing face information in (b) and (c) causes false positive detections (purple bboxes).
    }
    \label{fig:detection_vis1}
\end{figure}

Tab. \ref{tab:detection_results} reports performance drop in detection AP for different anonymization methods.
The no-anonymization baseline refers to RTMDet applied to the original images.
The AP does not reach 100 due to pose-based NMS in pseudo GT generation.
RTMDet has only bouning box-based NMS and its AP reaches 98.1 compared to pseudo GT.
Both black rectangle and DeepPrivacy2 reduce detection quality.
These methods deform the head region, often truncating bounding boxes around the head (Fig. \ref{fig:detection_vis2}).
The black rectangle always alters the head shape, but detection can still succeed if ears or hair are visible.
DeepPrivacy2 replaces the face but can fail completely, leading to missed detections, as shown in Fig. \ref{fig:detection_vis2}.
BLANKET achieves over 90 AP compared to the original detections.
The drop is caused by a different number of detections (both FPs and FNs) compared to the pseudo ground truth (Fig. \ref{fig:detection_vis1}).
Note that 90 AP is a very high result -- the current SOTA \cite{coDETR} on the COCO dataset is 66.0 AP.

Tab. \ref{tab:pose_results} reports pose estimation AP.
The second column includes all 17 COCO keypoints; the third excludes the five facial keypoints (ears, eyes, nose).
Both columns show evaluation on the pseudo GT bounding boxes for fair comparison.
The last column evaluates full in-the-wild performance: detection and pose estimation on anonymized images.

\begin{figure}[tb]
    \centering
    \includegraphics[width=0.49\linewidth,page=5]{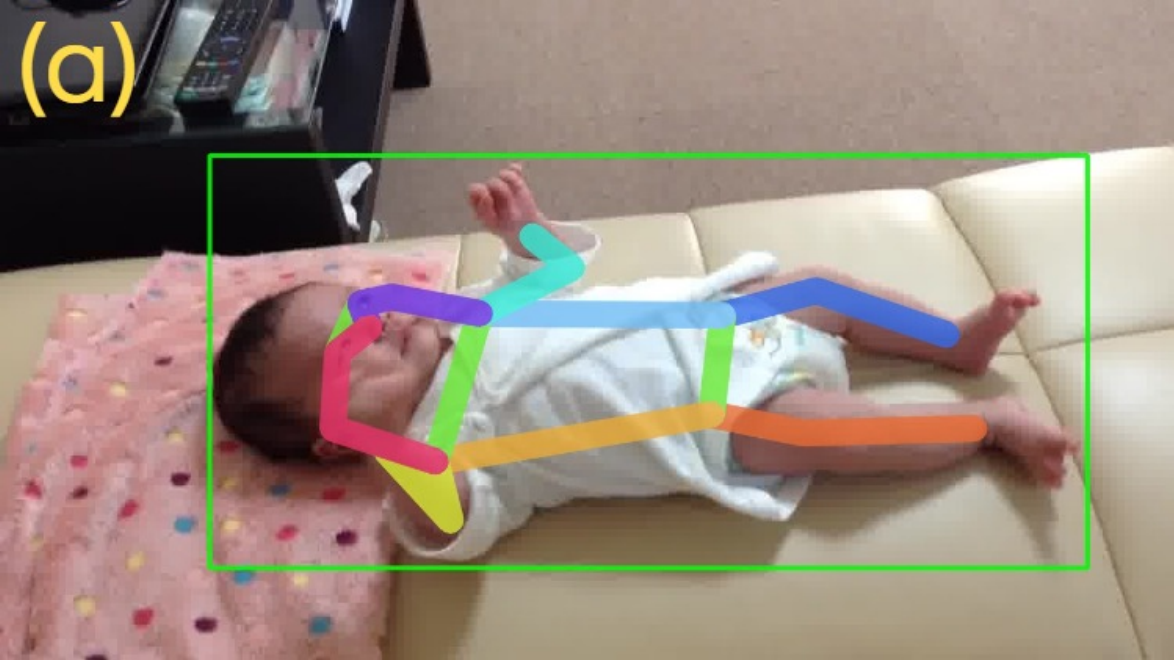}
    \hfill
    \includegraphics[width=0.49\linewidth,page=6]{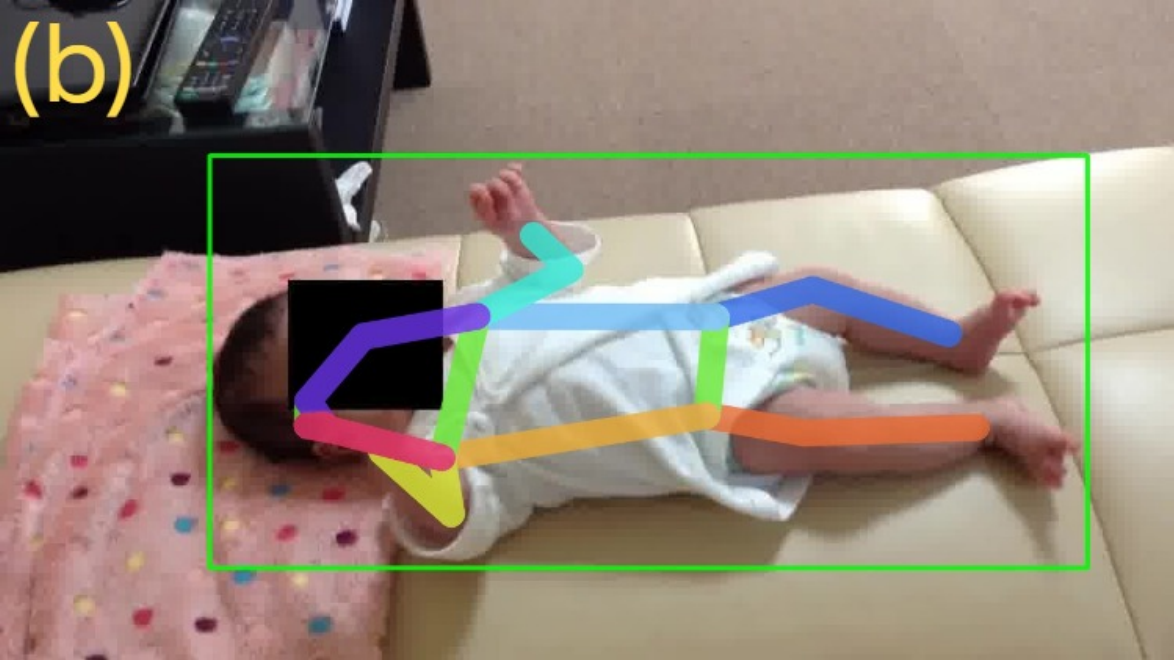}
    \vspace{-1.8ex}
    
    \includegraphics[width=0.49\linewidth,page=7]{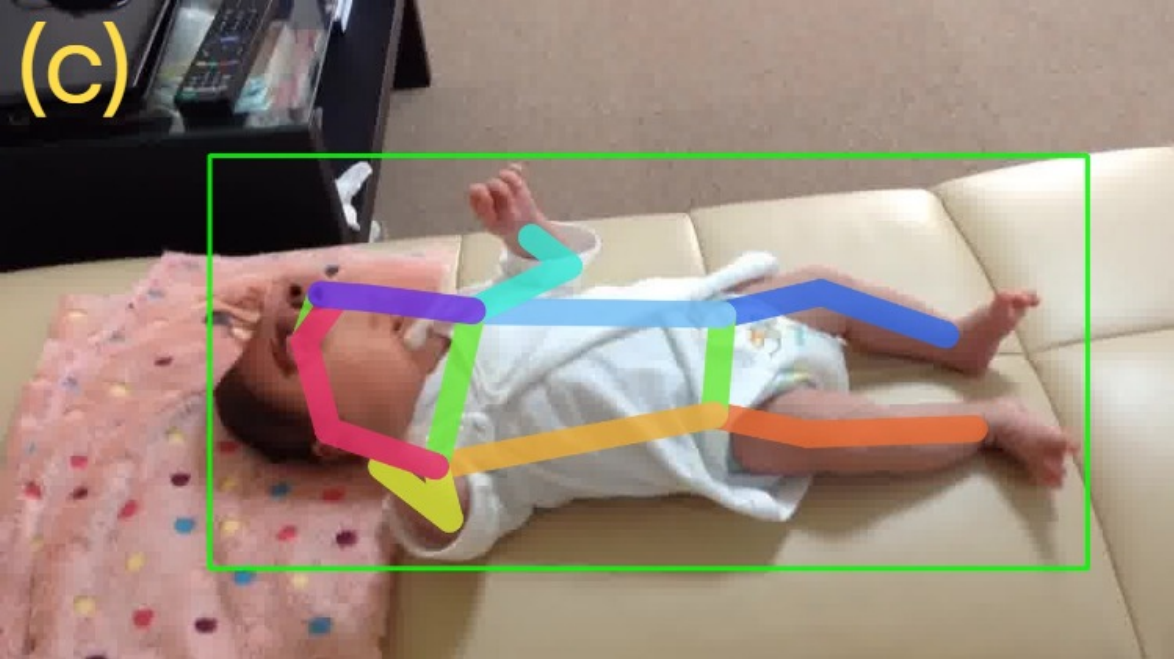}
    \hfill
    \includegraphics[width=0.49\linewidth,page=8]{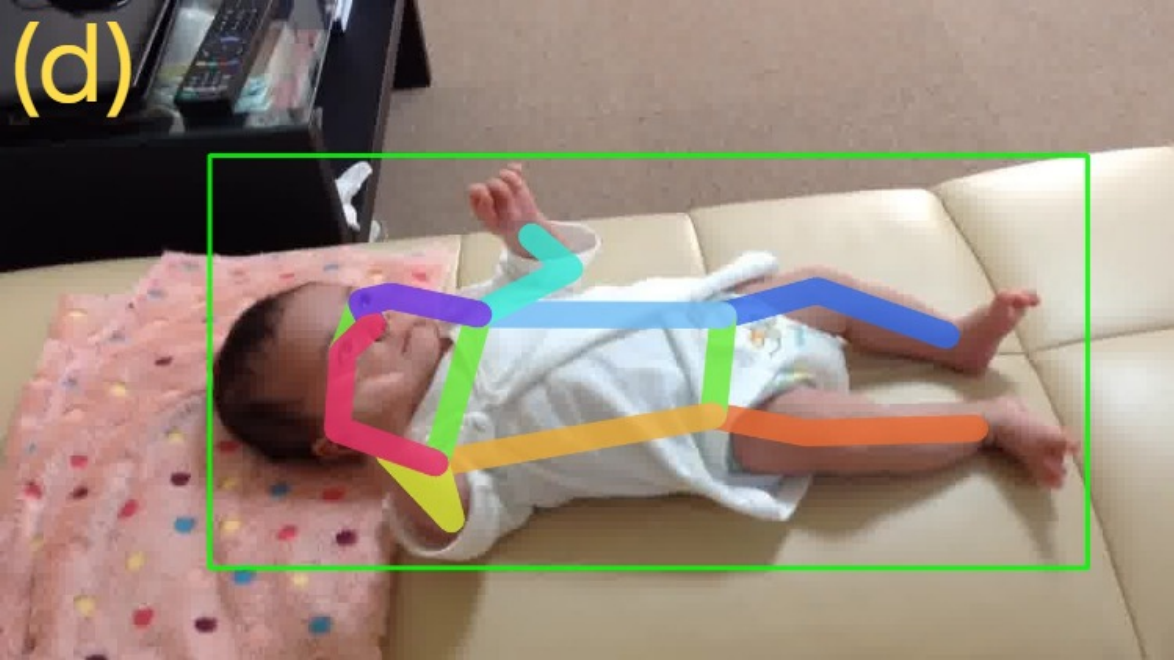}
    \caption{
    % Influence of face anonymization on human pose estimation.
    Anonymization methods hinder facial pose estimation.
    Original image (a), black rectangle anonymization (b), DeepPrivacy2 (c) and BLANKET (d) with their respective poses detected by ViTpose \cite{ViTPose}.
    Methods (b) and (c) dramatically hinder pose estimation on facial keypoints.
    }
    \label{fig:pose_vis1}
\end{figure}

\begin{figure}[tb]
    \centering
    \includegraphics[width=0.24\linewidth,page=1]{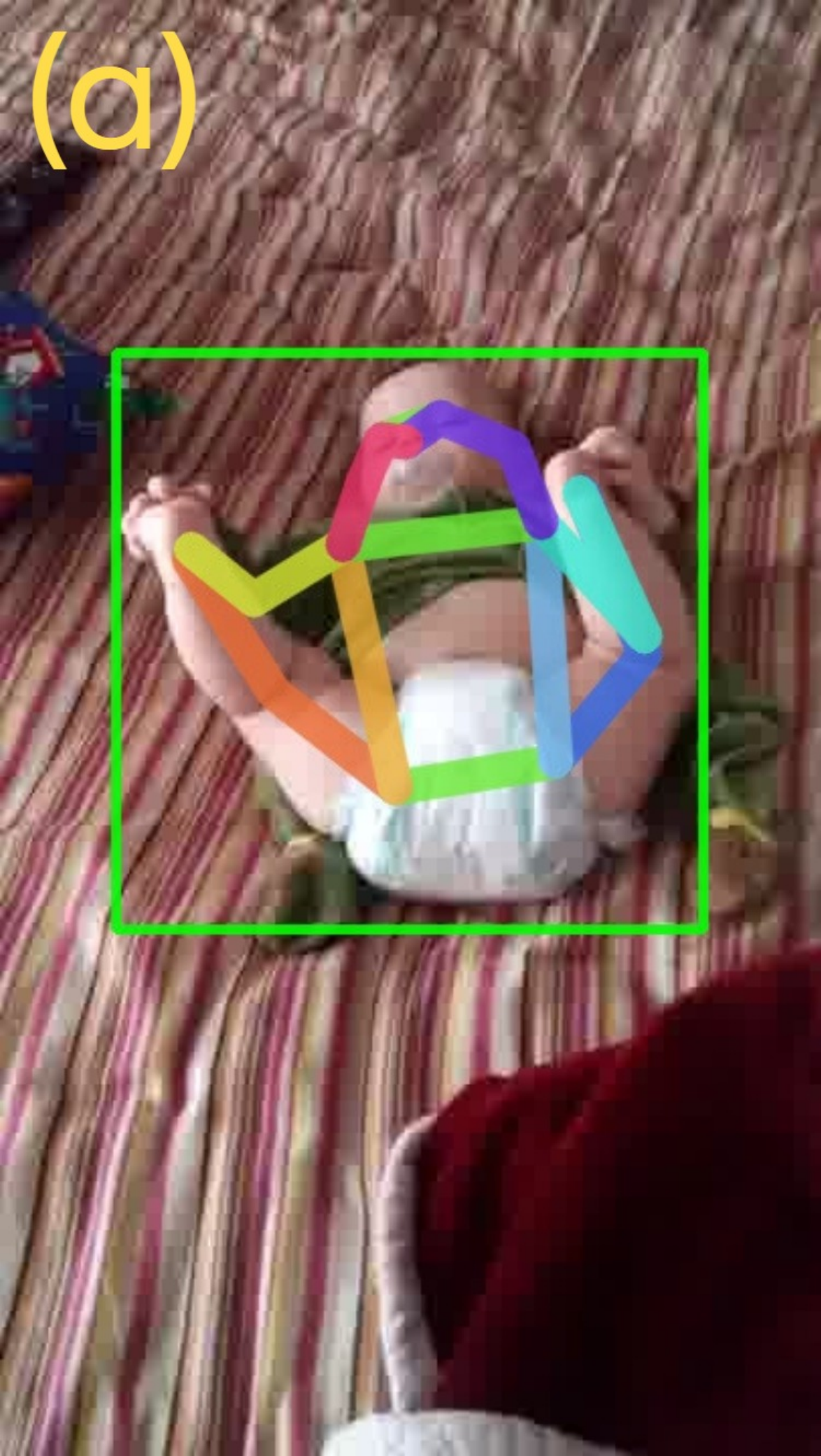}
    \hfill
    \includegraphics[width=0.24\linewidth,page=2]{images/Infants_anonymization_vis.pdf}
    \hfill
    \includegraphics[width=0.24\linewidth,page=3]{images/Infants_anonymization_vis.pdf}
    \hfill
    \includegraphics[width=0.24\linewidth,page=4]{images/Infants_anonymization_vis.pdf}
    \caption{
    % Influence of face anonymization on human pose estimation.
    Face is crucial for human pose estimation in uncommon poses.
    Original image (a), black rectangle anonymization (b), DeepPrivacy2 (c) and BLANKET (d) with their respective poses detected by ViTpose \cite{ViTPose}.
    Missing face information in (b) and (c) causes complete failure in uncommon poses.
    }
    \label{fig:pose_vis2}
\end{figure}

Images without anonymization achieve 100 AP by design, since they define the pseudo GT.
BLANKET yields the highest pose AP among anonymization methods.
DeepPrivacy preserves body keypoints but degrades performance on facial keypoints (Fig. \ref{fig:pose_vis1}).
The black rectangle degrades performance not only on facial keypoints but also on body keypoints.
Deformed head features in rare poses (e.g., infants in supine position) confuse the model, leading to limb mismatches or failure to understand the scene (Fig. \ref{fig:pose_vis2}).

In-the-wild performance corresponds to the combination of detection and pose estimation.
The black rectangle retains less than 20\% of the original AP, DeepPrivacy2 over 70\%, and our method over 90\%.

The analysis shows that detection contributes more to performance loss than pose estimation.
Publishing datasets with GT bounding boxes and anonymized faces would preserve nearly 98\% of pose estimation performance.

\subsection{User study} \label{sec:user_study}
We conducted a user study to assess the level of de-identification, and level of perceptual artifacts. Both of these respects are hard to measure automatically. The standard face recognition engine~\cite{ArcFace} is not reliable for infants. Perceptual artifacts, disturbing for human observers, are difficult to predict in the images automatically. We tried~\cite{zhang2023perceptual}, but it did not work. 

The questionair had two parts. The first part concerned the videos, the second part the images (frames). We had two questionairs with different instances. Each of the two contained 5 videos and 9 images. About 40 users filled our questionairs. The videos and images were chosen randomly (not cherry picked). 

For videos, we presented 3 short clips -- the original, and anonymization results by our method and by DeepPrivacy2. We did not reveal which of the method is ours and randomly shuffled the order of results. The question was ``Rate how much method A modified the identity of the face. Please focus on the face only -- not the hair, the body, clothes etc.'' The options were in range 1--5, where 1 meant ``identities are the same'' and 5 ``identities are completely different''. Another question ``Which video contains fewer artifacts?''

% \begin{figure}
% \centering
% \def\hght{0.36\linewidth}
% \tabcolsep=0pt
% \begin{tabular}{cc}
% \includegraphics[height=\hght]{images/user_study_id.pdf} &
% \includegraphics[height=\hght]{images/user_study-img.pdf} \\
% \footnotesize(a) & \footnotesize(b)
% \end{tabular}
% \caption{User study: (a) Histograms of subjective level of de-identification in range 1 (same identity) -- 5 (completely different identity). (b) Percentage of tested images for which users believe they are orginal (red), or artificially modified (blue) for three sources: Original footage, anonymization by our method, and by DeepPrivacy 2.}
% \label{fig:user_study}
% \end{figure}

\begin{figure}
\centering
\includegraphics[width=0.65\linewidth]{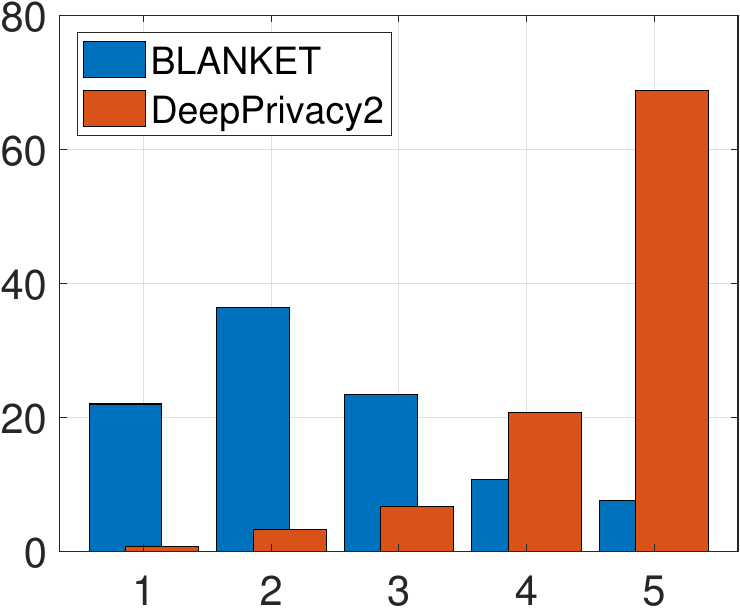} 
\caption{User study -- Histograms of subjective level of de-identification in range 1 (same identity) -- 5 (completely different identity).}
\label{fig:user_study_deindentification}
\end{figure}

\begin{figure}
\centering
\includegraphics[width=0.7\linewidth]{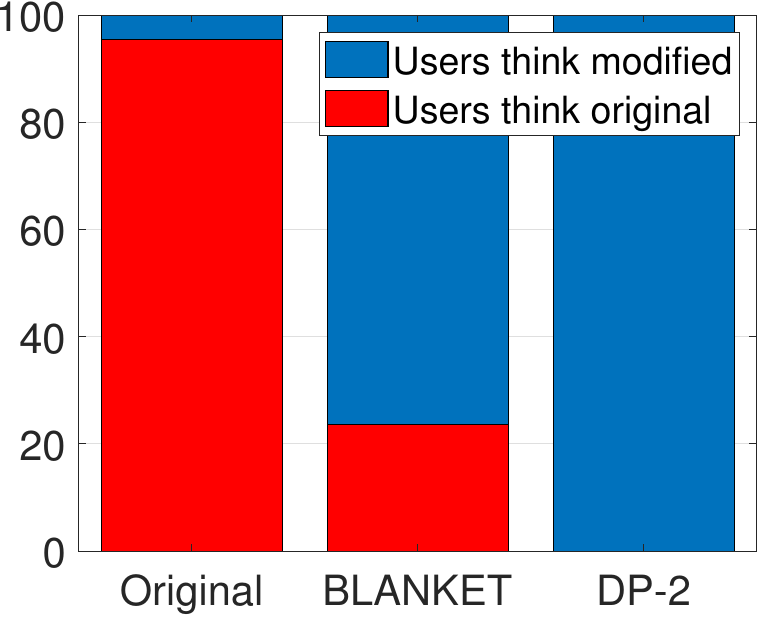} 
\caption{User study -- Percentage of tested images for which users believe they are orginal (red), or artificially modified (blue) for three sources: Original footage, anonymization by our method BLANKET, and by DeepPrivacy2.}
\label{fig:user_study_artifacts}
\end{figure}

The results are presented in Fig.~\ref{fig:user_study_deindentification}. It is seen that DeepPrivacy2 has higher level of de-identification, average scale is 4.5. This is apparent since the face is very often completely broken. Our method achieves an average scale 2.5. The identity was perceived to be altered, but often close to the original identity. We believe that this could be partially explained by the small resolution of the footage, since the method changes only the inner face. The difference is more obvious when the face is of higher resolution, where different eyes and other facial features appear clearly, as can be seen in our qualitative results. Users identified our anonymization, producing less artifacts than the competing method in 96\% of the cases. 

For images, we asked users, if they are able to spot the anonymization traces. So we randomly shuffled 9 frames from the original, anonymized by our method, and anonymized by DeepPrivacy2. Each source has 3 images. The question was binary: ``The face in the image above is either (a) the original unmodified face or (b) a modified (edited, artificially manipulated) face''. 

The results are shown in Fig.~\ref{fig:user_study_artifacts}. We can see the distribution of user responses. In total, frames of original recordings were identified in 96\% of the cases. The frames produced by our anonymization method were confused with the original frames in 24\% of the cases, demonstrating that the method does not produce obvious disturbing artifacts. Moreover, most of our users were computer vision students who know where to look for subtle artifacts. Nobody confused the frames produced by DeepPrivacy2 in any case with the original, which confirms that the method does not generalize well for infant subjects. 

\subsection{Qualitative results} \label{sec:qualitative}

\def\txt#1&#2&{\setbox\tmpbox\hbox{#2}\hskip-3pt%
    \vbox to\ht\tmpbox{\vfill\hbox{#1}\vfill}&#2&}

\begin{figure*}
    \def\wdth{0.3695\linewidth}
    \def\twdth{0.118}
    \centering
    \tabcolsep=1pt
    \begin{tabular}{c ccccc}     
        %\txt\rotatebox{90}{\footnotesize{DeepPrivacy2 \hspace{20pt} BLANKET \hspace{27pt} Original}} & \hspace{-1.5ex}
        \rotatebox{90}{\begin{tabular}{p{\twdth\linewidth}p{\twdth\linewidth}p{\twdth\linewidth}}
            \centering \footnotesize{DeepPrivacy2} & \centering \footnotesize{BLANKET} & \centering \footnotesize{Original}
        \end{tabular}} & \hspace{-1ex}
        \includegraphics[trim=0mm 0mm 70mm 0mm,clip,height=\wdth]{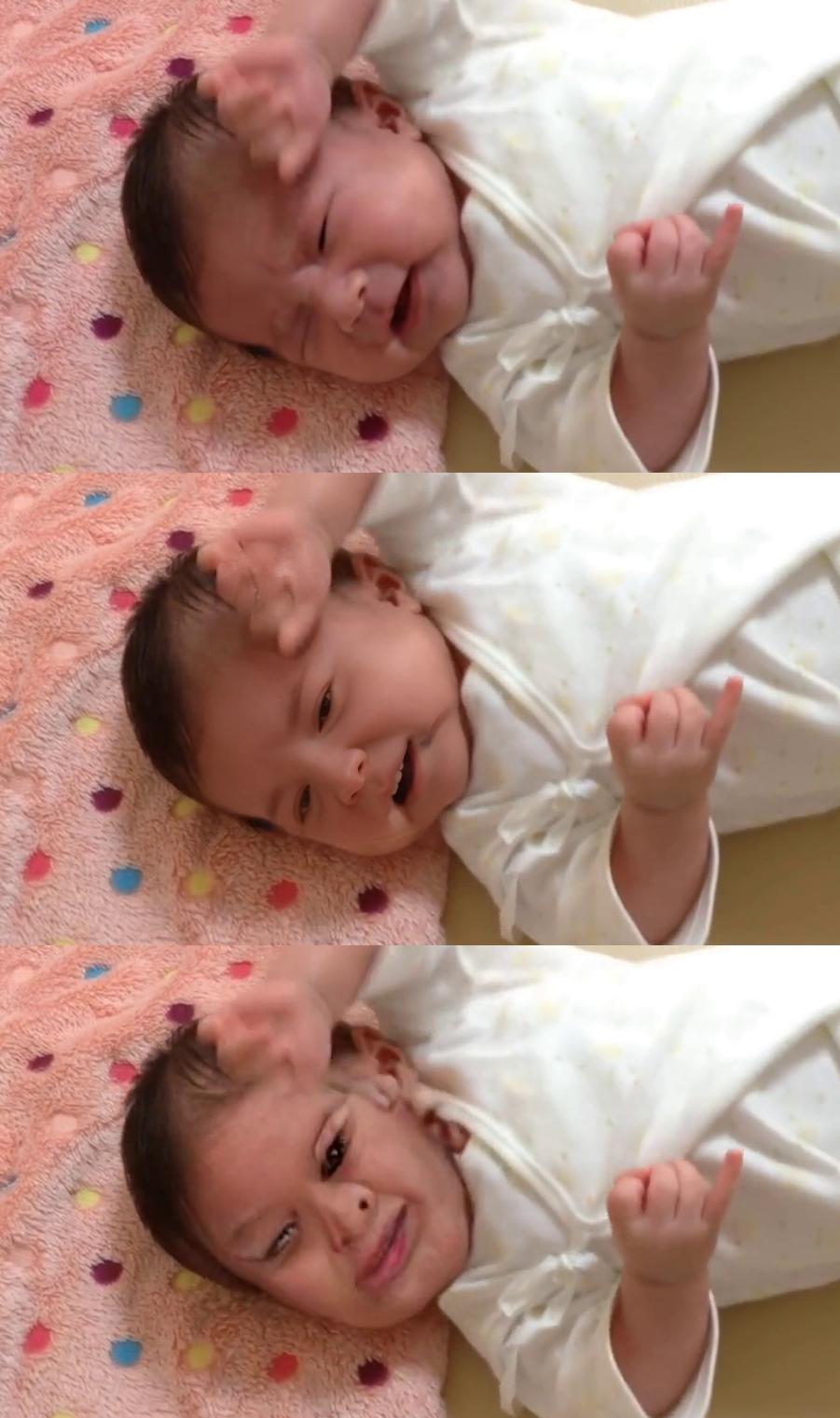} &
        \includegraphics[trim=30mm 0mm 0mm 0mm,clip,height=\wdth]{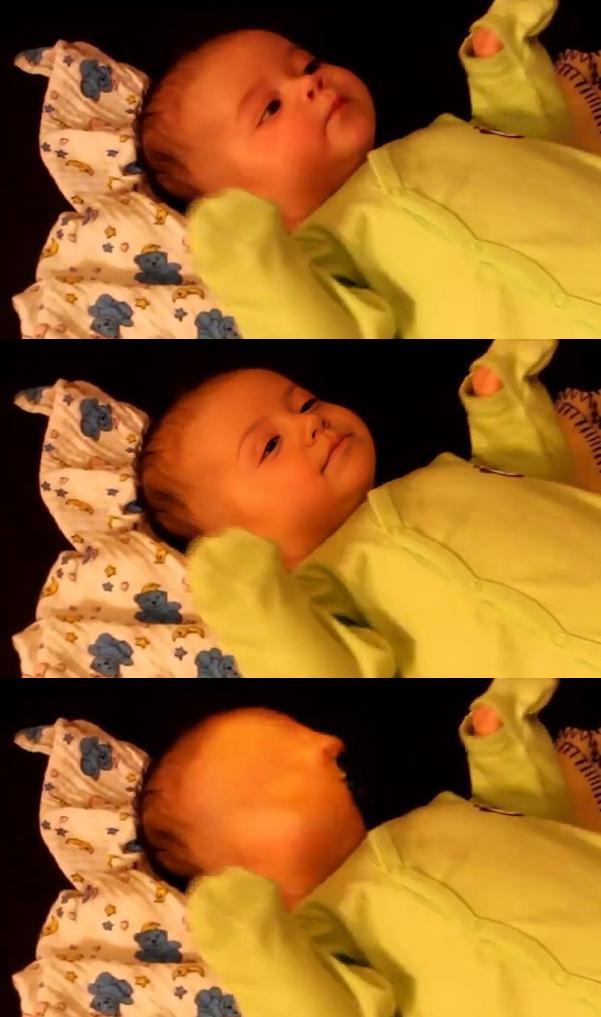} &
        \includegraphics[trim=0mm 0mm 30mm 0mm,clip,height=\wdth]{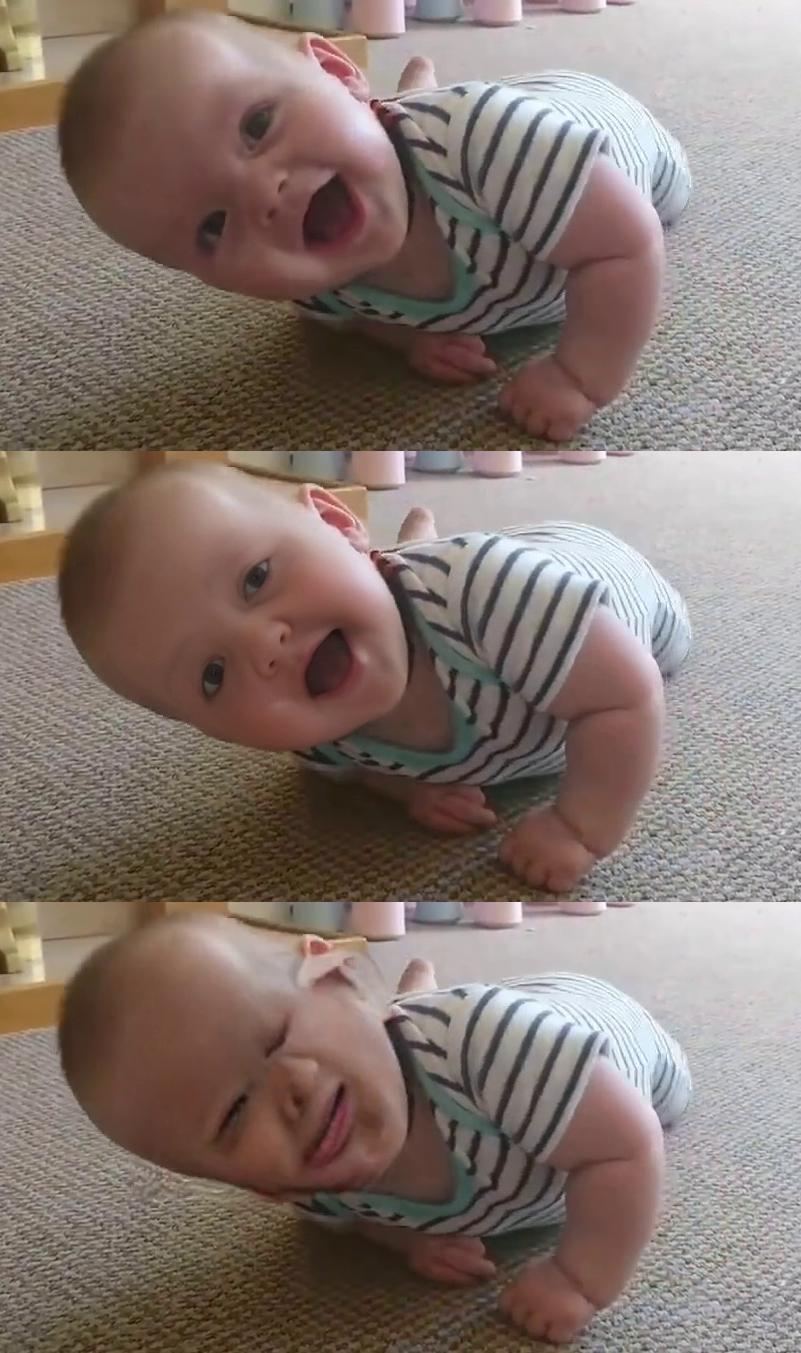}& 
        \includegraphics[trim=30mm 0mm 20mm 0mm,clip,height=\wdth]{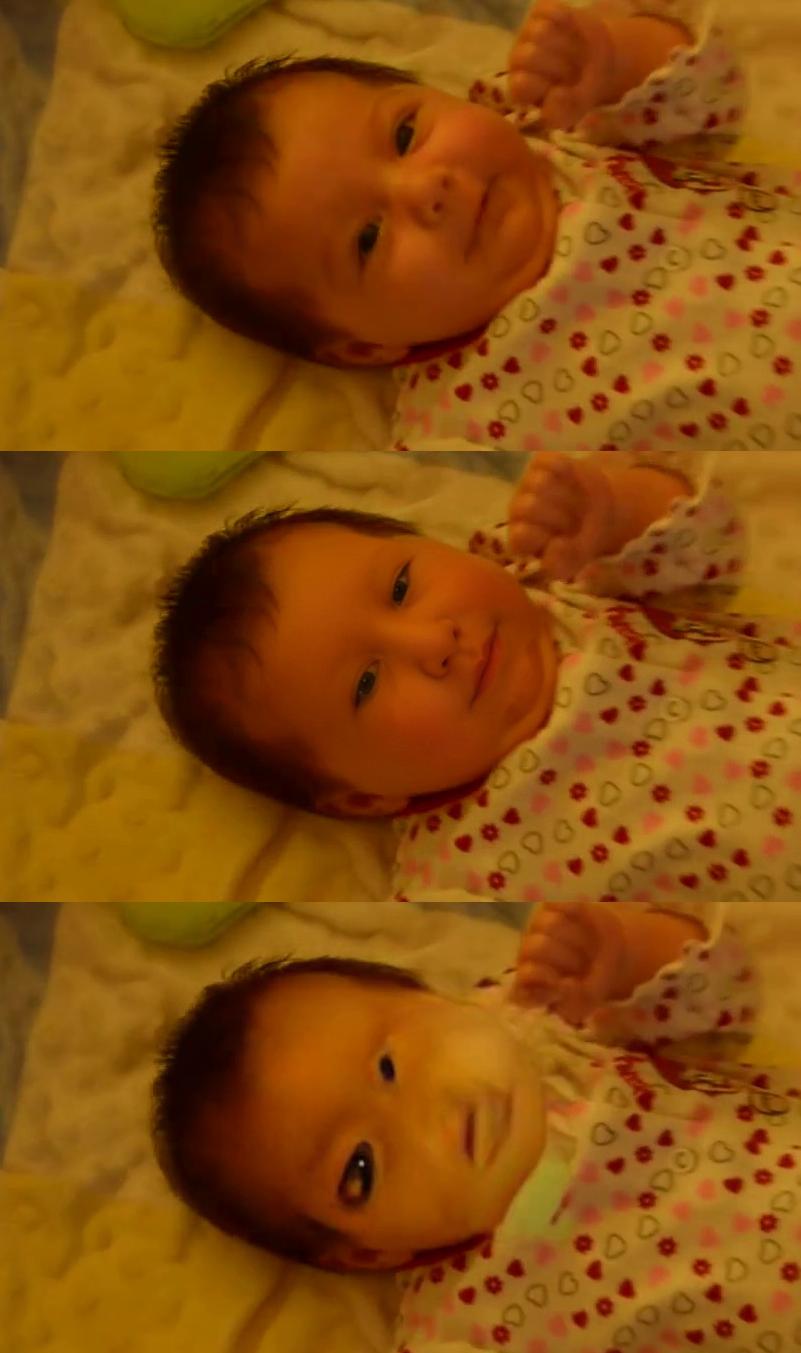} &
        \includegraphics[height=\wdth]{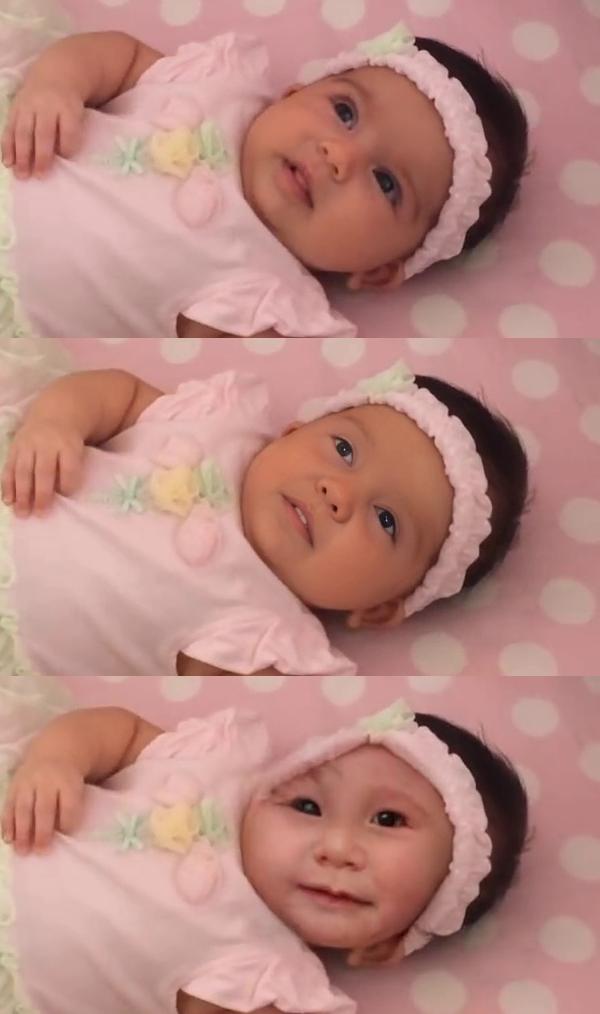} \\[2ex]
    \end{tabular}

    \def\wdth{0.42\linewidth}
    \def\twdth{0.136}
    \centering
    \tabcolsep=1pt
    \begin{tabular}{c ccccc}   
        \rotatebox{90}{\begin{tabular}{p{\twdth\linewidth}p{\twdth\linewidth}p{\twdth\linewidth}}
            \centering \footnotesize{DeepPrivacy2} & \centering \footnotesize{BLANKET} & \centering \footnotesize{Original}
        \end{tabular}} & \hspace{-1ex}
        \includegraphics[trim=10mm 0mm 0mm 0mm,clip,height=\wdth]{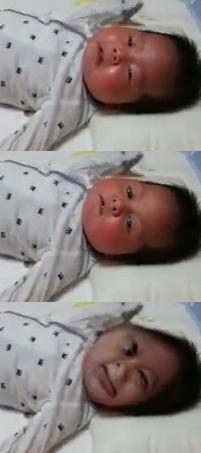} &
        \includegraphics[trim=0mm 0mm 10mm 0mm,clip,height=\wdth]{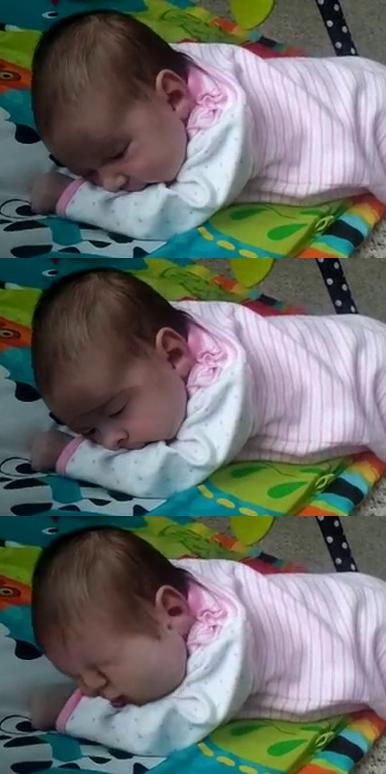} &
        \includegraphics[trim=0mm 0mm 30mm 0mm,clip,height=\wdth]{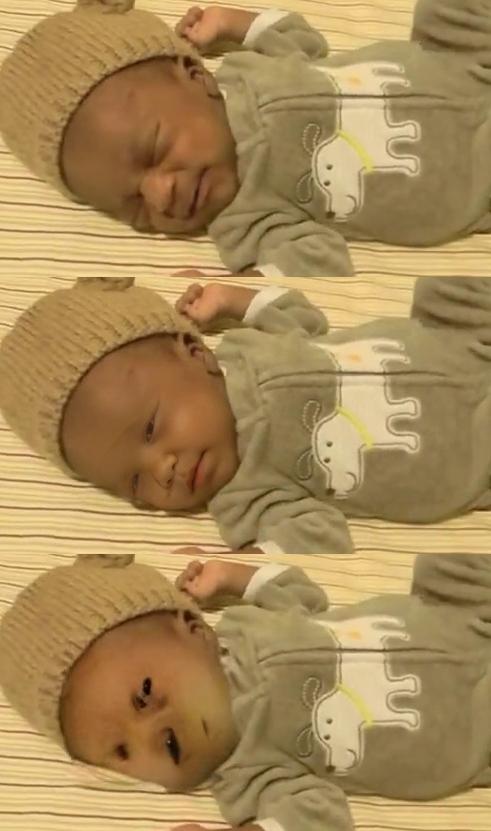} & 
        \includegraphics[trim=60mm 0mm 0mm 0mm,clip,height=\wdth]{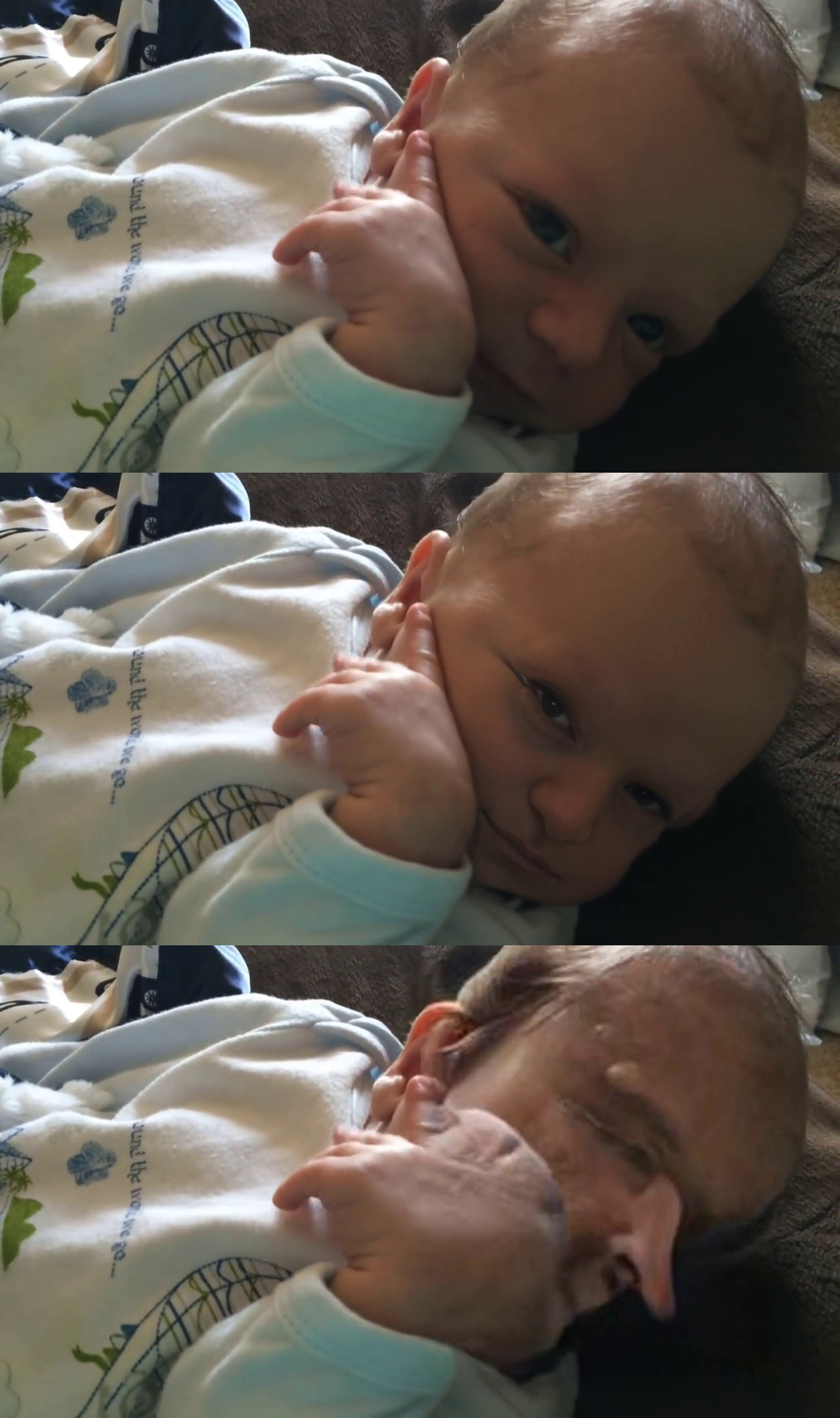} &
        \includegraphics[trim=10mm 0mm 20mm 0mm,clip,height=\wdth]{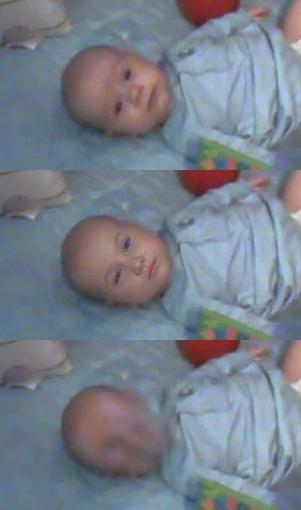} 
    \end{tabular}    
    
    \caption{Qualitative results. Frames of the original videos are compared with anonymization by our method BLANKET and by DeepPrivacy2. Please, zoom in to see the details of the facial features. See supplementary videos at \url{ https://github.com/ctu-vras/blanket-infant-face-anonym}.}
    \label{fig:qr1}
\end{figure*}

In Figs.~\ref{fig:teaser} and~\ref{fig:qr1}, we show several qualitative results. The figures depict a comparison of original frames, anonymized by our method and by DeepPrivacy2. Our method can handle challenging conditions, such as occlusions and unusual viewpoints (off-the-plane rotations), while mostly preserving the facial expression of the original. We observed that occasionally the model does not generate the eyes fully closed. We recommend zooming in to see that the identity has changed. See our supplementary video for an idea on the stability and temporal consistency of the anonymization of the video recordings. 

Unlike our method, DeepPrivacy2 suffers from severe artifacts, both spatial and temporal. Although the method is promising for anonymizing adult subjects, it is unusable for infants. The anonymized faces are not matching or very distorted. Anonymization changes the face dramatically from frame to frame. The first problem is probably caused by out-of-domain distribution, since the GAN-based method had difficulties to generate faces of infants and there seem to be problems with feature alignment of infants face, which cause non-matching results. The latter problem is that the method applies the anonymization to every frame independently, which causes temporal inconsistency.

\section{Conclusion, Discussion and Future Work} \label{sec:conclusion}

%We presented a seamless anonymization method that was tuned and extensively tested for infant video recordings. The method alters the identity of a subject and preserves attributes of the original video, including facial expressions. We conclude that the proposed method has a negligible impact on human pose estimation, which is an important downstream task for, e.g., behavior analysis. The experiments show that our method BLANKET generates temporally consistent results, unlike the competing method DeepPrivacy2, which turned out to be unsuitable for anonymizing infant videos. 

We presented a seamless anonymization method tuned and extensively tested on infant video recordings. The method alters a subject’s identity while preserving original video attributes, including facial expressions. We conclude that the proposed method has a negligible impact on human pose estimation, an important downstream task for, e.g., behavior analysis. Experiments show that our method, BLANKET, generates temporally consistent results, unlike the competing method, DeepPrivacy2, which proved unsuitable for anonymizing infant videos.

%A precondition for face anonymization is the correct detection of the face in the image. If the face detector fails or detects another face in the frame, the face to be anonymized will not be modified. This is addressed provisionally by obscuring the whole frame. In the future, interpolation from neighboring anonymized images can be used. 

Anonymization relies on accurate face detection; if it fails or detects another face in the frame, the target may remain unmodified. Currently, such cases are handled by obscuring the entire frame, with future improvements exploring interpolation from adjacent anonymized frames.

%The limitation of our method is that sometimes the level of de-identification perceived by the users seemed low, so the novel identity resembles the original. This could be partly due to the small resolution of the faces, and one has to zoom in to see the detailed features. On the other hand, we only change the inside of the face, and keep the facial landmark intact, which inherently contains certain identity information. It would be challenging to modify the shape of the entire face. Another option would be to generate identities that are maximally dissimilar to the original, which is hard to do automatically, since face recognition engines do not work well on infants, which we tested. In our future work, we will investigate this direction and test other metrics to find dissimilar subjects, but still compatible, with minimum artifacts. In a similar spirit as in~\cite{Subrtova-2024}, we will investigate the options to detect and correct disturbing perceptual artifacts. 

A limitation of our method is that users occasionally perceived the de-identification as insufficient, with the anonymized face resembling the original. This may stem from the low resolution of infant faces, where detailed features require zooming. Additionally, the method alters only the facial interior, retaining landmarks that inherently carry identity cues. Modifying the full facial structure is complex. An alternative is to generate identities maximally dissimilar to the original, though difficult to automate, especially since, as we verified, face recognition engines perform poorly on infants. Future work will explore metrics for selecting dissimilar yet compatible identities with minimal artifacts. Following~\cite{Subrtova-2024}, we will also examine methods for detecting and correcting perceptual artifacts.

From this perspective, it is advantageous for privacy and security that contemporary face recognition engines do not reliably work on infants, making it unlikely for individuals to be easily identified in large-scale datasets, unlike adult subjects. However, with sufficient data, face recognition models can be fine-tuned. The challenge lies in the scarcity of available data and the ethical concerns surrounding the creation of such datasets containing raw imagery. Subjectively, newborns and babies share a high degree of similarity, making identification challenging even for human observers. As a result, relatively small modifications to the face further complicate re-identification.

%\red{The code will be made publicly available after the acceptance of the paper. It will appear here: \url{ https://github.com/ctu-vras/blanket-infant-face-anonym}}.

% Challenges
% \begin{itemize}
%     \item hands in front of face
%     \item eyes closed are a problem (hard to preserve)
%     \item if infant is not upright
%     \item out of plane rotation - not en face, z profilu
%     \item occlusions, not complete face in the picture
%     \item black screen or black rectangle. detector failed? face fusion....
%     \item automatic evaluation of perceptual artifacts (citovat Eliho a nas clanek)
%     \item future work - degree of anonymization; face recognition does not work on infants 
% \end{itemize}

\section*{Acknowledgment}
We would like to thank Adéla Šubrtová who was part of the discussions in the initial stages of the project. %She also put together the dataset of videos sourced from Youtube. 

%\section*{References}
\bibliographystyle{IEEEtran}
\bibliography{bibliography}

\end{document}